\documentclass{hld2026} 

\usepackage{amsmath,amsfonts,bm}

\def\eqref#1{equation~\ref{#1}}

\def\1{\bm{1}}

\def\vq{{\bm{q}}}

\DeclareMathAlphabet{\mathsfit}{\encodingdefault}{\sfdefault}{m}{sl}
\SetMathAlphabet{\mathsfit}{bold}{\encodingdefault}{\sfdefault}{bx}{n}

\def\sD{{\mathbb{D}}}

\newcommand{\R}{\mathbb{R}}

\usepackage{placeins}
\usepackage{booktabs}
\usepackage{amsmath}
\usepackage{tikz}
\usepackage{amssymb}
\usepackage{float} 
\usepackage{caption}

\title[Scaling Laws for Grid-Based Approximate Nearest Neighbor Search in High Dimensions]{Scaling Laws for Grid-Based Approximate Nearest Neighbor Search in High Dimensions}

\hldauthor{%
\Name{Matthew J. Liu} \Email{matthewliu@berkeley.edu}\\
\addr University of California, Berkeley
\AND 
\Name{Wei Hang Zheng} \Email{weihang\_zheng@berkeley.edu}\\
\addr University of California, Berkeley 
\AND
\Name{Vidhan Purohit} \Email{viddyp1804@gmail.com}\\
\addr University of Toronto, St. George
\AND
\Name{Siqi Xie} \Email{siqixie@alumni.cmu.edu}\\
\addr Independent Researcher
\AND
\Name{Chieh-En Li} \Email{chiehenli@gmail.com}\\
\addr Independent Researcher
\AND
\Name{Jerry Li} \Email{li.zhiyi.jerry@gmail.com}\\
\addr University of Waterloo
\AND
\Name{Noah Flynn} \Email{noahflynn@berkeley.edu}\\
\addr University of California, Berkeley
}

\begin{document}

\maketitle

\begin{abstract}
Grid-based approaches to approximate nearest neighbor (ANN) search have been absent from modern scaling analyses. We present a systematic characterization of a multiprobe grid algorithm with respect to dataset size $N$ and dimensionality $d$. Our experiments reveal a previously unreported $d$-scaling crossover on the GloVe embedding family, in which multiprobe grid search maintains an approximately constant dimensional scaling exponent while other graph-, tree-, and partitioning-based methods exhibit degrading throughput. The advantage comes with near-linear query scaling in $N$, but also with lower indexing cost than competing ANN methods. Our results suggest that grid-based methods such as multiprobe grid may be competitive in rebuild-heavy or high-dimensional settings where indexing cost and dimensional robustness dictate performance. More broadly, recent work has formalized self-attention as an ANN operation. Thus, the $N$- and $d$-scaling properties of ANN algorithms may guide cost analysis of efficient transformer architectures. Code is available at: \url{https://github.com/weiz345/MultiProbeANN}. 
\end{abstract}


\section{Introduction}

Approximate nearest neighbor (ANN) search is central to modern machine learning systems operating on large, high-dimensional datasets \cite{darrell_nearest-neighbor_2005}. Beyond its classical role in retrieval, ANN has emerged as a computational primitive in transformer architectures. Recent studies formalize self-attention as an ANN operation over token embeddings, enabling sub-quadratic approximations to full attention \cite{haris_knn_2025, liu_fast_2025, kang_attention_2025}. As empirical scaling laws guide transformer design \cite{kaplan_scaling_2020}, the scaling behavior of ANN algorithms is increasingly relevant to the development of more efficient transformer architectures. In particular, scaling dataset size $N$ stresses candidate filtering and indexing efficiency, while scaling dimensionality $d$ exacerbates the curse of dimensionality and degrades the effectiveness of geometric pruning. As a result, ANN methods including graph-, tree-, and partitioning-based approaches exhibit regime-dependent trade-offs in which no method dominates universally \cite{xiao_enhancing_2024, iff_benchmarking_2026}. 

Grid-based methods (a variant of partitioning approaches) played a foundational role in early ANN theory \cite{indyk_approximate_1998, chan_approximate_1997}, but remain undercharacterized relative to graph- and tree-based methods in modern scaling analyses (see Appendix~\ref{sec:related_works_appendix} for Related Work). In this work, we characterize $N$ and $d$ scaling relationships for a multiprobe grid algorithm (overview in Section \ref{sec:theory}) that decouples cell selection from $d$, where cell selection denotes the query-time procedure that identifies which grid cells supply candidate vectors. This decoupling is achieved by performing cell selection in a PCA-reduced subspace $\mathbb{R}^m$ while re-ranking candidates in $\mathbb{R}^d$. We show that while multiprobe grid exhibits near-linear scaling with $N$, its $d$-scaling behavior remains favorable as dimensionality increases. This work situates grid-based methods within the broader ANN design space and highlights regimes in which they offer competitive trade-offs.

\section{Theoretical scaling model for multiprobe grid search}
\label{sec:theory}
We provide a theoretical overview for the multiprobe grid algorithm, deriving a closed-form relationship between query cost and recall. Appendix~\ref{sec:proof} contains the full discussion and Figure \ref{fig:figureD1} provides an overview schematic for the algorithm.
\paragraph{Cost model.} After PCA projection to $\mathbb{R}^m$, the space is partitioned into $G^m$ cells with $N/G^m$ points each. A query $q$ probes a total of $1 \le n_{\mathrm{probe}} \le 2^m$ cells, consisting of the home cell $c_h$ (defined in Appendix~\ref{app:setup}) and up to $2^m-1$ neighboring cells ordered ascending by wall distance $w_i^2$ (Appendix~\ref{app:costmon}). Each probed cell contributes, on average, $N/G^m$ candidates to the cost:
\begin{equation}
\label{eq:cost}
\mathrm{cost} \;=\; 1/\mathrm{QPS} \;=\; K \cdot n_{\mathrm{probe}} \cdot \frac{N}{G^m}.
\end{equation}
\paragraph{Recall model.} Let $P_i = \Pr[x^* \in c_i]$ be the probability that the true nearest neighbor lies in the $i$-th probed cell. Under the uniform distribution assumption, $P_i$ decreases monotonically in $w_i^2$, and we adopt the exponential approximation (Appendices~\ref{app:mem} and~\ref{app:logqps}).
\begin{equation}
\label{eq:pdf}
P_i \;\approx\; P_h \, e^{-\mu w_i^2}, \qquad P_h := P_0,\ \mu > 0
\end{equation}
\paragraph{Closed form.} Using a linear approximation $\mathbb{E}[w_i^2] \approx \theta i$ (Appendix~\ref{app:expected}) on the mean gap yields
\begin{equation}
\label{eq:recall}
R(n_{\mathrm{probe}}) \;=\; P_h \sum_{i=0}^{n_{\mathrm{probe}}-1} \Delta^i \;=\; P_h \cdot \frac{1 - \Delta^{n_{\mathrm{probe}}}}{1 - \Delta},  \Delta := e^{-\mu\theta} \in (0,1)
\end{equation}
\paragraph{Log-linearity.} Solving~\eqref{eq:recall} for $n_{\mathrm{probe}}$, substituting into~\eqref{eq:cost}, and inverting gives
\begin{equation}
\label{eq:qps_exact}
\mathrm{QPS}(R) \;=\; \frac{G^m}{KN}\cdot\frac{|\ln\Delta|}{-\ln(1 - R/R_{\max})}.
\end{equation}
Taking logs, we arrive at $\log \mathrm{QPS}(R) \approx \log(\mathrm{const}') - \log R$. Thus, our framework predicts a log-linear relationship between QPS and recall, arising from the exponential decay of nearest-neighbor membership probability across probed cells and the linear growth of candidate set size with $n_{\mathrm{probe}}$.

\section{Empirical scaling laws}

We evaluate our multiprobe grid algorithm against four baselines representing major ANN families: Voyager (graph-based) \cite{noauthor_spotifyvoyager_2026}, PyNNDescent (graph-based)~\cite{mcinnes_lmcinnespynndescent_2026}, Annoy (tree-based) \cite{noauthor_spotifyannoy_2026}, and FAISS-IVF (quantization-based partitioning)~\cite{noauthor_facebookresearchfaiss_2026}. Algorithms are evaluated using the \texttt{ann-benchmarks} framework~\cite{aumuller_ann-benchmarks_2020, aumuller_reproducibility_2021}, executing each method in a dedicated Docker container to enforce single-CPU isolation. The baseline implementations use highly optimized C++ implementations, while our multiprobe grid implementation is a Python-based proof-of-concept. Consequently, the measured QPS for multiprobe grid likely understates the throughput attainable with a comparably optimized implementation. Nonetheless, we expect that the relative scaling trends reported here remain informative, given that multiprobe grid's per-query time is primarily spent in NumPy/BLAS (see profiling study in Appendix~\ref{sec:methods}). Baseline algorithms use their established \texttt{ann-benchmarks} parameter sweeps, which represent well-explored, community-validated search spaces; the multiprobe grid algorithm required bespoke NSGA-II tuning to identify competitive configurations (Appendix~\ref{sec:methods} for implementation details).

\subsection{Pareto fronts reveal a log-linear throughput-recall relationship for multiprobe grid}
\label{sec:pareto}
\begin{figure}[h]
    \centering
    \includegraphics[width=0.82\linewidth]{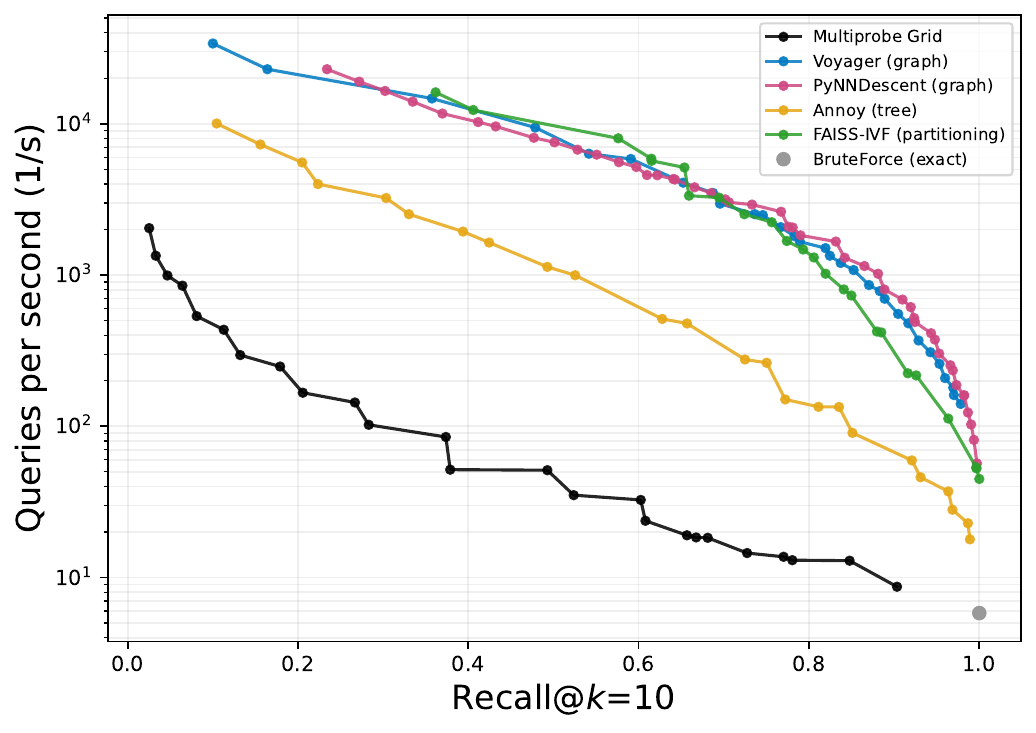} 
    \caption{Pareto fronts for GloVe-200-angular ($d=200$, $N=1.18\times10^6$).}
    \label{fig:figure1} 
\end{figure}

Figure \ref{fig:figure1} shows Pareto fronts on GloVe-200-angular ($N=1.18\times10^6$ points) \cite{pennington_glove_2014, aumuller_ann-benchmarks_2020, aumuller_reproducibility_2021} for all five algorithms. Consistent with Section \ref{sec:theory}, multiprobe grid shows that $\log(\mathrm{QPS})$ decreases linearly with increasing recall, indicating that performance is determined by grid geometry. At recall@$k$=10 $> 0.9$, multiprobe's throughput converges toward brute-force: achieving high recall requires either coarse PCA projections ($m{=}2$) that produce densely-populated cells, or exhaustive multiprobing at higher $m$, resulting in ranking a large fraction of the dataset. We show individual Pareto fronts at each subsampled $N$ for GloVe-200 and $d$ (GloVe-25, 50, 100, 200) in Figures \ref{fig:figureD2}--\ref{fig:figureD3}. 

\subsection{$N$-scaling: multiprobe grid scales near-linearly with dataset size}
\subsubsection{GloVe dataset (angular)}
\begin{figure}[h]
    \centering
    \includegraphics[width=0.95\linewidth]{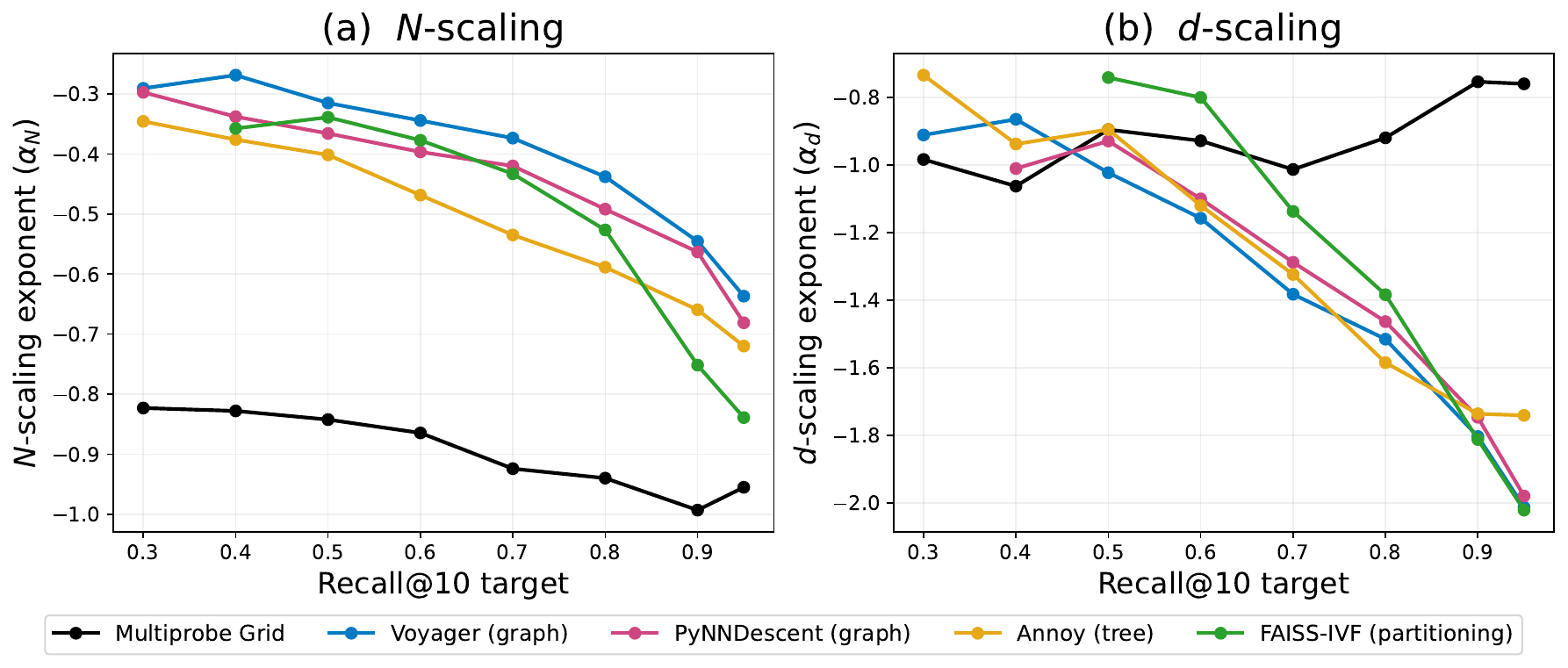} 
    \caption{ \textbf{Scaling exponents vs. recall@$k$ =10 for all five algorithms}. \textbf{(a)} $N$-scaling exponent $\alpha_N$. Datasets: GloVe-200-angular ($d=200)$ subsampled at varying $N$. \textbf{(b)} $d$-scaling exponent $\alpha_d$. Datasets: GloVe-25-, 50-, 100-, and 200-angular ($N=1.18\times10^6$). } 
    \label{fig:figure2} 
\end{figure}

Figure \ref{fig:figure2}a shows how the $N$-scaling exponent $\alpha_N$ varies with recall across all five ANN algorithms. At recall@$k$=10 $= 0.80$, multiprobe grid exhibits $\alpha_N = -0.94$ ($R^2 = 1.00$), indicating near-linear degradation of QPS with dataset size. In contrast, we observe sublinear scaling for the baselines, ranging from $\alpha_N = -0.44$ to $-0.59$. Figure~\ref{fig:figureD4} shows the $\log_{10}(\mathrm{QPS})$ vs. $\log(N)$ plots at each target recall@$k$=10 used to derive the exponents in Figure \ref{fig:figure2}a. 

The near-linear $N$-scaling of multiprobe is mechanistically expected: for fixed grid parameters $(m, G)$, the candidate count per cell grows as $N / G^m$, and re-ranking cost is linear in candidate count. As recall targets increase, all algorithms' exponents trend toward $\alpha_N = -1$, reflecting how perfect recall requires exhaustive search. Multiprobe's exponent is already near its asymptotic value across the full recall range, while baseline methods degrade more rapidly at higher recalls.

\subsubsection{SIFT-128 dataset (Euclidean)}
To investigate whether the observed $N$-scaling relationships generalize beyond word embeddings, we repeat the analysis on SIFT-128-euclidean (image descriptors, $d=128$) \cite{jegou_product_2011}. The log-linear Pareto behavior of multiprobe grid (Figure~\ref{fig:figureD5}) and relative trends in $\alpha_N$ are preserved (Figures~\ref{fig:figureD6}--\ref{fig:figureD7}). At recall@$k$=10 $= 0.80$, multiprobe grid exhibits a scaling exponent $\alpha_N = -0.83$ ($R^2 = 1.00$), while baseline algorithms fall between $-0.27$ and $-0.39$. The consistency across data modality (images vs. words) and similarity metric (Euclidean vs.\ angular) suggests that $N$-scaling in multiprobe grid is intrinsic to the algorithm rather than dataset-specific.

\subsection{$d$-scaling: multiprobe grid is increasingly competitive for higher dimensional data}
\label{sec:dscaling}
Figure~\ref{fig:figure2}b shows how the $d$-scaling exponent $\alpha_d$ varies with recall, revealing a non-monotonic crossover between multiprobe grid and the other four algorithms. As recall increases beyond 0.7, $\alpha_d$ of all algorithms besides multiprobe grid steepen dramatically, reflecting lower throughput with increasing dimensionality. Meanwhile, $\alpha_d$ for multiprobe grid remains relatively flat. The contrast is notable given that prior empirical studies, including our own in Figure \ref{fig:figure2}a, typically find $N$-scaling exponents to be remarkably consistent across algorithm families \citep{sun_scaling_2025}. The $d$-scaling crossover subverts this universality - the same algorithm that exhibits the least favorable $N$-scaling simultaneously exhibits the most favorable $d$-scaling at high recall. Figure~\ref{fig:figureD8} shows the $\log_{10}(\mathrm{QPS})$ vs. $\log(d)$ plots at each target recall@$k$=10 used to derive the exponents in Figure \ref{fig:figure2}b. 

The crossover in $\alpha_d$ arises from how each algorithm interacts with $d$. Graph-based methods (Voyager, PyNNDescent) construct and traverse proximity graphs in the full $d$-dimensional space. At high recall, accurate retrieval requires exploring larger neighborhoods and more backtracking. Similarly, tree-based (Annoy) and quantization-based partitioning (FAISS-IVF) methods operate on the raw $d$-dimensional vectors, where pruning effectiveness degrades as $d$ grows. In contrast, multiprobe grid performs cell selection in a PCA subspace of dimensionality $m \ll d$. Although PCA retains a smaller fraction of total variance as $d$ increases (Figure \ref{fig:figureD9}), the Pareto-optimal $(m, G)$ adapts with both $d$ and the target recall rather than remaining fixed (Appendix Table~\ref{tab:tableE1}). At higher $d$, fewer total cells $G^m$ are favored, concentrating more candidates per probed cell to preserve recall. Because re-ranking grows only linearly in candidate count, the query cost scales less aggressively with $d$ than for methods that traverse the full $d$-dimensional space. We note that our $d$-scaling characterization is bounded by $d=200$, the maximum dimensionality available in the GloVe family (the only $d$-varying dataset family provided in \texttt{ann-benchmarks}). While this range demonstrates the $\alpha_d$ crossover, extending our analysis to higher-$d$ regimes (e.g., $d \ge 512$ for modern transformer embeddings) is a key direction for future work (Section \ref{conclusion}).

\subsection{Total cost analysis reveals competitive regimes for multiprobe grid}
\label{sec:build}
Figure~\ref{fig:figureD10} shows build time scaling with $N$ for the Pareto-optimal configuration achieving recall@$k$ = 10 $\approx 0.80$. Multiprobe grid and FAISS-IVF are fastest across all $N$. At $N=1.18\times10^6$, multiprobe grid builds its index in 4–36 sec depending on the selected hyperparameter configuration (4 sec for the smallest config, $m{=}2, G{=}4$; 36 sec for the largest, $m{=}7, G{=}7$). We measure 206 sec for FAISS-IVF, 333 sec for Annoy, 500 sec for PyNNDescent, and 1,569 sec for Voyager. Multiprobe grid indexing requires only PCA fitting, cell assignment, and BFS precomputation, whereas baselines rely on data-dependent operations such as graph insertion by repeated nearest-neighbor queries (Voyager, PyNNDescent).

Recent work by Sun et al. established power-law scaling relationships for different ANN methods like brute-force, partitioning-based, and graph-based \cite{sun_scaling_2025}. The study proposed a framework for the total cost $J$ of ANN search that is a function of query-time compute, indexing, and storage costs: $J(N) = f_I \cdot I(N) + f_C \cdot C(N) + S(N)$, where $I$, $C$, and $S$ denote indexing, compute, and storage costs, and $f_I$, $f_C$ their respective frequency of occurrence. We adapt the cost framework with one substitution, reporting memory footprint $M(N)$ in place of $S(N)$, because \texttt{ann-benchmarks} measures process RSS during indexing rather than on-disk storage size. Table~\ref{tab:table1} shows empirical scaling exponents at recall@$k$=10 $= 0.80$ on GloVe-200-angular. $C(N)$ corresponds directly to the values shown in Figure~\ref{fig:figure2}a for recall@$k$=10 $= 0.80$. $I(N)$ and $M(N)$ are measured directly from the Pareto-optimal configuration at each $N$ (Figures~\ref{fig:figureD10} and \ref{fig:figureD11}). At $N=1.18\times10^6$, multiprobe grid builds its recall@$k$=10 $= 0.80$ index $\sim$190$\times$ faster than Voyager (8.4 sec vs.\ 1,569 sec), but exhibits per-query latency approximately 120$\times$ slower. Thus, multiprobe grid achieves lower total cost when the rebuild-to-query ratio $f_I/f_C$ is sufficiently high: approximately one index rebuild per 2,600--20,400 queries depending on the baseline algorithm (see Appendix \ref{sec:crossover}). This regime may include applications in recommendation systems and retrieval-augmented generation. 

\begin{table}[h]
\centering

\caption{Scaling exponents at recall@$k$=10 $= 0.80$ (GloVe-200, angular; $R^2$ in parentheses).}

\label{tab:table1}
\small

\begin{tabular}{llccc}
\toprule
Algorithm & Family & $C(N)$ & $I(N)$ & $M(N)$ \\
\midrule
\textbf{Multiprobe Grid} & \textbf{grid} & $N^{0.94}$ (1.00) &
$N^{1.20}$ (0.82) & $N^{0.73}$ (0.96) \\
Voyager (HNSW) & graph & $N^{0.44}$ (0.92) & $N^{1.27}$ (0.97) &
$N^{0.94}$ (1.00) \\
PyNNDescent & graph & $N^{0.49}$ (0.94) & $N^{0.74}$ (0.90) &
$N^{0.52}$ (0.92) \\
Annoy & tree & $N^{0.59}$ (0.99) & $N^{1.09}$ (0.87) &
$N^{0.87}$ (0.93) \\
FAISS-IVF & partition & $N^{0.53}$ (0.96) & $N^{2.00}$ (0.84) &
$N^{0.81}$ (0.99) \\
\bottomrule
\end{tabular}
\end{table}

\subsection{Implications for high-dimensional learning systems}
\label{conclusion}
Building on the formalization of attention as ANN search \cite{liu_fast_2025, haris_knn_2025, kang_attention_2025}, our scaling characterization maps dataset size $N$ to context length, native dimensionality $d$ to the attention head dimension ($d_{\text{head}}$, typically $64$ to $128$, within the range characterized here), and indexing cost to the incremental updates as new tokens extend the KV-cache. In particular, low-cost indexing methods are favored when the ratio of index updates to queries is high (e.g., during incremental KV-cache growth). This regime may favor multiprobe grid, as appending a token to the grid requires only a single cell assignment rather than a full rebuild. Thus, the nominal rebuild cost $I(N)$ is a substantially cheaper per-insert cost; characterizing this per-insert cost and its interaction with the precomputed BFS fallback is left to future work. An additional, empirical question for future work is whether the $\alpha_d$ crossover persists for Key vector distributions in transformers, which we anticipate exhibit greater non-uniformity and heavier-tailed structures than GloVe. More broadly, our results underscore that ANN method selection should consider $N$- and $d$-scaling relationships, total cost $J$, and operational frequencies such as rebuild rate.

\bibliography{references}
\newpage

\appendix

\renewcommand{\thefigure}{A\arabic{figure}}
\setcounter{figure}{0}

\section{Related work} 
\label{sec:related_works_appendix}

Approximate nearest neighbor (ANN) search retrieves high-quality candidate sets approximating true nearest neighbors, trading exactness for speed \cite{indyk_approximate_1998, jayaram_subramanya_diskann_2019}. Indexing strategies include tree-based, hash-based, quantization-based, and graph-based methods \cite{xiao_enhancing_2024}. Among these, hashing and quantization are often grouped under a broader partitioning-based framework, as both divide the search space into discrete regions to prune candidates \cite{sun_scaling_2025}. We adopt this framing in our work.

Grid-based methods, a variant of partitioning-based methods, were explored in early theoretical studies on ANN. For example, translated grids were used to mitigate exponential dependence on dimensionality and remove logarithmic factors in batched query settings \cite{chan_approximate_1997}. Subsequent work, such as the bucketing method, achieved constant-time hash evaluations independent of dataset size \cite{indyk_approximate_1998}. These approaches are notable for their conceptual simplicity and minimal indexing overhead. As ANN research shifted toward large-scale, high-dimensional datasets, practical focus moved toward more flexible methods, including Locality-Sensitive Hashing (LSH) \cite{gionis_similarity_1999, datar_locality-sensitive_2004, jafari_experimental_2021}, k-d trees \cite{bentley_multidimensional_1975, arya_optimal_1998, silpa-anan_optimised_2008}, and Production Quantization \cite{jegou_product_2011}. As a result, grid-based methods became relatively underexplored in modern ANN systems, despite offering a distinct set of structural and theoretical properties.

In addition to partitioning-based methods, modern ANN systems frequently rely on graph-based indices, particularly Hierarchical Navigable Small World (HNSW), which achieve strong empirical performance by traversing multi-layer proximity graphs \cite{malkov_efficient_2020}. However, graph-based methods introduce distinct operational challenges. For example, they can exhibit capacity-limited failure, where insufficient exploration leads to abrupt degradation in neighbor quality \cite{cooper_capacity-limited_2026}. They are also sensitive to dynamic updates, as repeated insertions and deletions can produce unreachable nodes and degrade recall over time \cite{xiao_enhancing_2024}. More broadly, these limitations highlight the complex trade-offs between query efficiency, index construction, and robustness that characterize modern ANN methods. 

Beyond algorithmic design, recent work has emphasized empirical evaluation and scaling behavior as the primary lens for understanding ANN performance. Standardized benchmarks such as \texttt{ann-benchmarks} evaluate methods across datasets, reporting recall–latency tradeoffs under fixed resource budgets \cite{aumuller_ann-benchmarks_2020, aumuller_reproducibility_2021}. Benchmarking studies have helped establish empirical baselines across a wide range of regimes, with HNSW and Facebook AI Similarity Search (FAISS) often exhibiting state-of-the-art performance. To better generalize benchmarks, studies also seek to evaluate how performance scales with dataset size $N$ and search budget. For example, Sun et al. established power-law scaling relationships for partitioning- and graph-based methods \cite{sun_scaling_2025}. The study proposed a framework for the total cost $J$ of ANN search that is a function of query-time compute, indexing, and storage costs: $J(N) = f_I \cdot I(N) + f_C \cdot C(N) + S(N)$, where $I$, $C$, and $S$ denote indexing, compute, and storage costs, and $f_I$, $f_C$ their respective frequency of occurrence. We adapt this cost framework in Section \ref{sec:build}.

While benchmarking and scaling studies have significantly improved our understanding of ANN performance, they primarily focus on graph-based, tree-based, and adaptive partitioning methods. Grid-based approaches have remained comparatively underexplored despite their unique properties in the ANN design space, such as structural simplicity, low indexing cost, and predictable query behavior. We anticipated that grid-based methods may therefore exhibit different performance trade-offs from the other, more characterized methods - particularly in regimes where indexing cost, rebuild frequency, or robustness to dimensionality are dominant concerns. Thus, we revisit grid-based ANN methods in this work, aiming to expand understanding of how different ANN design principles translate into distinct scaling behaviors.

\section{Derivation of the Multiprobe Grid Scaling Model}
\label{sec:proof}
We derive a probabilistic upper-bound approximation for the query cost--recall relationship of multiprobe grid search, averaged over many queries. The derivation assumes uniform query and data distribution at the cell scale and does not account for rank distortion introduced by the PCA projection; extending to PCA-induced distortion is left for future work. Throughout, we use the same notation as the main text: $N$ data points in $\mathbb{R}^d$, projected via PCA to $\mathbb{R}^m$ and partitioned into a uniform $G^m$ grid with cell side length $C$.

\subsection{Probe Setup and Cell Offsets}
\label{app:setup}

Given a query $\vq \in \mathbb{R}^d$ and its PCA projection $\vq' = (q'_1, \ldots, q'_m)$, let $c_h$ denote the home cell. For each dimension $j$, define the binary offset
\[
o_j \;=\; \begin{cases} +1 & q'_j \ge \text{midpoint of } c_h \text{ in dim. } j, \\ -1 & \text{otherwise.} \end{cases}
\]
Because the position of $\vq'$ within $c_h$ determines the sign per dimension, the set of candidate cell displacements is the Cartesian product of the $m$ binary choices, giving $2^m$ offset vectors. Removing the all-zero vector (the home cell) leaves $2^m - 1$ neighboring cells, rather than the $3^m - 1$ full-neighborhood count. This orthant pruning is employed to prune the number of cells checked as $m$ increases.

\paragraph{Two-dimensional example.} For $m = 2$ with $q'$ above the midpoint in both dimensions ($o = (+1, +1)$), the non-home offset vectors are $\{(0,1), (1,0), (1,1)\}$ in Cartesian choice notation, corresponding to the horizontal, vertical, and diagonal neighbors in the upper-right orthant.

\subsection{Wall distance ordering and cost model}
\label{app:costmon}
For each candidate cell, define the squared wall distance
\begin{equation}
\label{eq:walldist}
w_i^2 \;=\; \sum_{j \,:\, o_j \ne 0} (q'_j - \mathrm{wall}_j)^2,
\end{equation}
where $\mathrm{wall}_j$ is the boundary of $c_h$ shared with the candidate cell in dimension $j$. The $2^m - 1$ values are sorted ascending as $(w_0^2, \ldots, w_{2^m - 2}^2)$, and at query time only the first $n_{\mathrm{probe}} \le 2^m - 1$ cells in this order are probed. Candidates from all probed cells are re-ranked in the native $d$-space.

Each probed cell contributes on average $N/G^m$ points to the candidate set, and ranking is linear in the candidate count. Thus
\begin{equation}
\label{eq:cost_app}
\mathrm{cost} \;=\; 1/\mathrm{QPS} \;=\; K \cdot n_{\mathrm{probe}} \cdot \frac{N}{G^m},
\end{equation}
where $K > 0$ absorbs per-candidate constants (distance computations, comparisons, bookkeeping).

\subsection{Recall as a sum of cell-membership probabilities}
\label{app:mem}
Let $x^*$ denote the true nearest neighbor of $q$ and define the cell-membership probability $P_i = \Pr[x^* \in c_i]$. Because probed cells are disjoint,
\[
R(n_{\mathrm{probe}}) \;=\; \sum_{i=0}^{n_{\mathrm{probe}}-1} P_i.
\]
Under uniform query/data distribution at the cell scale, $P_i$ decreases monotonically with $w_i^2$; since the $w_i^2$ are sorted ascending, $P_0 \ge P_1 \ge \cdots \ge P_{n_{\mathrm{probe}}}$. We adopt the exponential fall-off
\begin{equation}
\label{eq:pdf_app}
P_i \;\approx\; P_h \, e^{-\mu w_i^2}, \qquad P_h := P_0, \ \mu > 0,
\end{equation}
as a mean-field approximation.

\subsection{Expected wall distance via offset geometry}
\label{app:expected}

To obtain an analytical form, we replace the cell-specific $w_i^2$ with its expected value under uniform queries. From~\eqref{eq:walldist}, $w_i^2 = \sum_{j : o_j \ne 0} \varepsilon_j^2$ where $\varepsilon_j := q'_j - \mathrm{wall}_j$. Under the uniform-distribution assumption, $\varepsilon_j \sim U(0, C/2)$ (the query sits in a given half of $c_h$ per dimension), giving
\[
\mathbb{E}[\varepsilon_j^2] \;=\; \int_0^{C/2} x^2 \cdot \frac{2}{C} \, dx \;=\; \frac{(C/2)^2}{3} \;=:\; L.
\]
Let $A_i$ denote the set of nonzero offset coordinates of the $i$-th cell. By linearity,
\begin{equation}
\mathbb{E}[w_i^2] \;=\; |A_i| \cdot L.
\end{equation}
For each $j \in \{1, \ldots, m\}$, there are $\binom{m}{j}$ offset vectors with exactly $j$ nonzero entries, all sharing $\mathbb{E}[w_i^2] = jL$. Sorting the $2^m - 1$ cells in ascending order of expected wall distance produces a step function that takes value $jL$ on a plateau of length $\binom{m}{j}$, for $j = 1, \ldots, m$.

\paragraph{Example: $m = 3$.} There are $\binom{3}{1} = 3$ face-adjacent cells, reachable by the offset vectors (0,0,1), (0,1,0) and (1,0,0) with $\mathbb{E}[w^2] = L$, $\binom{3}{2} = 3$ edge-adjacent cells with offset vectors (0,1,1), (1,0,1) and, (1,1,0) with $\mathbb{E}[w^2] = 2L$, and $\binom{3}{3} = 1$ corner cell with offset vector (1,1,1) with $\mathbb{E}[w^2] = 3L$, totaling $2^3 - 1 = 7$.

\paragraph{Arithmetic mean gap (telescoping).} Define $\mathbb{E}[w_{-1}^2] := 0$ for the home cell. The arithmetic mean of first differences is
\[
\theta \;=\; \frac{1}{2^m - 1}\sum_{i=0}^{2^m - 2} \bigl(\mathbb{E}[w_i^2] - \mathbb{E}[w_{i-1}^2]\bigr).
\]
The sum telescopes: within each plateau the increment is $0$, and at each of the $m$ transitions between consecutive plateaus the increment is $L$. Therefore
\begin{equation}
\theta \;=\; \frac{m \cdot L}{2^m - 1} \;=\; \frac{m \,(C/2)^2}{3 \,(2^m - 1)}.
\end{equation}
For $m = 3$, $\theta = 3L/7 = (C/2)^2/7$. We approximate the step function linearly as
\begin{equation}
\mathbb{E}[w_i^2] \;\approx\; \theta \, i.
\end{equation}

\subsection{Geometric closed form}

Substituting $\mathbb{E}[w_i^2] \approx \theta i$ into~\eqref{eq:pdf_app} with $\Delta := e^{-\mu \theta} \in (0, 1)$, we obtain $P_i \approx P_h \Delta^i$, and the recall becomes a geometric series:
\begin{equation}
\label{eq:recall_app}
R(n_{\mathrm{probe}}) \;=\; P_h \sum_{i=0}^{n_{\mathrm{probe}}-1} \Delta^i \;=\; P_h \cdot \frac{1 - \Delta^{n_{\mathrm{probe}}}} {1 - \Delta}.
\end{equation}
The achievable recall range is bounded by
\[
R_{\min} \;=\; P_h \quad (n_{\mathrm{probe}} = 1), \qquad R_{\max} \;=\; P_h \cdot \frac{1 - \Delta^{2^m}}{1 - \Delta} \quad (n_{\mathrm{probe}} = 2^m).
\]
For $\Delta \in (0,1)$ and moderately large $m$, $\Delta^{2^m} \to 0$ very rapidly, so
\begin{equation}
\label{eq:rmax_app}
R_{\max} \;\approx\; \frac{P_h}{1 - \Delta}
\qquad\Longleftrightarrow\qquad
\frac{1}{R_{\max}} \;\approx\; \frac{1 - \Delta}{P_h}.
\end{equation}

\subsection{Log-QPS derivation: from geometric closed form to exponential}
\label{app:logqps}

Starting from the geometric closed form~\eqref{eq:recall_app} and solving for $n_{\mathrm{probe}}$,
\[
\Delta^{n_{\mathrm{probe}}} \;=\; 1 - \frac{R(1 - \Delta)}{P_h}, \qquad n_{\mathrm{probe}}(R) \;=\; \frac{\ln(1 - R/R_{\max})}{\ln \Delta}.
\]
Using the approximation $(1 - \Delta)/P_h \approx 1/R_{\max}$ from~\eqref{eq:rmax_app},
\begin{equation}
\label{eq:nprobe_R_app}
n_{\mathrm{probe}}(R) \;=\; \frac{\ln(1 - R/R_{\max})}{\ln \Delta} - 1.
\end{equation}
Substituting~\eqref{eq:nprobe_R_app} into the cost formula~\eqref{eq:cost_app} and factoring out $\ln \Delta$ from the numerator,
\begin{equation}
\label{eq:one_over_qps}
\frac{1}{\mathrm{QPS}} \;=\; \frac{KN}{G^m}\cdot\frac{\ln(1 - R/R_{\max}) - \ln \Delta}{\ln \Delta}.
\end{equation}
Both $\ln(1 - R/R_{\max})$ and $\ln \Delta$ are negative (since $R/R_{\max}, \Delta \in (0,1)$), so we may rewrite~\eqref{eq:one_over_qps} with all signs made explicit:
\begin{equation}
\label{eq:qps_explicit}
\mathrm{QPS}(R) \;=\; \frac{|\ln \Delta|}{K N / G^m} \cdot \frac{1}{-\ln(1 - R/R_{\max}) + \ln \Delta}.
\end{equation}

\paragraph{Taking the log.} To expose the exponential structure, we take the logarithm of both sides of~\eqref{eq:qps_explicit}:
\begin{equation}
\label{eq:log_qps_raw}
\log \mathrm{QPS}(R) \;=\; \log\!\Big(\tfrac{|\ln\Delta| G^m}{KN}\Big) \;-\; \log\!\Big(-\ln(1 - R/R_{\max}) + \ln \Delta\Big).
\end{equation}
The first term is constant in $R$; all the $R$-dependence lives in the second term.

\paragraph{Taylor expansion of the inner log.} Let $u := R/R_{\max} \in (0, 1)$. The function $-\ln(1 - u) = \ln\!\bigl(1/(1-u)\bigr)$ admits the convergent Taylor series
\[
-\ln(1 - u) \;=\; u + \tfrac{u^2}{2} + \tfrac{u^3}{3} + \cdots
\]
on $(0, 1)$. Because $u < 1$, the higher-order terms are dominated by the linear term, and we truncate at first order:
\begin{equation}
\label{eq:taylor_app}
-\ln(1 - R/R_{\max}) \;\approx\; \frac{R}{R_{\max}}.
\end{equation}

The first-order truncation in~\eqref{eq:taylor_app} has relative error $1 - u/[-\ln(1-u)]$, which is approximately $28\%$ at $u = 0.5$, $50\%$ at $u \approx 0.8$, and diverges as $u \to 1$. The closed form derived below is therefore a local approximation, accurate for $u$ bounded away from $1$. As we discuss in Appendix~\ref{app:validity}, this is one of two approximations whose validity degrades in the high-recall regime; the empirical Pareto front in Figure~\ref{fig:figure1} nonetheless follows the predicted log-linear shape over most of the operating range, suggesting that the structural prediction survives quantitative slack in the underlying assumptions. Substituting~\eqref{eq:taylor_app} into~\eqref{eq:log_qps_raw},
\begin{equation}
\label{eq:log_qps_taylor}
\log \mathrm{QPS}(R) \;\approx\; \log(\mathrm{const}) \;-\; \log\!\Big(R/R_{\max} + \ln \Delta\Big).
\end{equation}

\paragraph{Mean-field approximation.} We now make the mean-field argument rigorous. The derivation rests on the following assumption:

\begin{quote}
\textbf{Assumption (mean-field).} The query points are uniformly distributed within the home cell, and the dataset points are drawn from a distribution that is approximately uniform at the scale of a single grid cell.
\end{quote}

Three consequences follow. First, the wall distances $w_i^2$ are expectations taken over the uniform query position distribution within the cell, so $\theta$ is the mean inter-neighbor gap averaged over all possible query locations (Appendix~\ref{app:expected}). Second, the decay parameter $\mu$ characterizes the rate at which the NN-membership probability falls off with wall distance, and is itself determined by the local point density at the cell scale. Lastly, since the point distribution is approximately uniform at the cell scale, the cell-membership probability mass $P_i = P_h e^{-\mu w_i^2}$ does not vary sharply between adjacent cells. That is, neighboring cells have similar density to the home cell, so the decay is gradual. This implies
\begin{equation}
\label{eq:meanfield_regime}
\mu\theta \;\ll\; 1
\qquad\Longleftrightarrow\qquad
\Delta = e^{-\mu\theta} \approx 1
\qquad\Longleftrightarrow\qquad
|\ln\Delta| \;\ll\; 1.
\end{equation}

This assumption is equivalent to requiring that the grid cell size be small relative to the scale at which the data distribution varies, which is the standard regime underlying any local approximation in partition-based ANN. It is also the regime in which the grid index is well-calibrated: if cells were so large that density varied significantly within them, the grid partition itself would be a poor indexing structure regardless of multiprobe.

\paragraph{Practical domain of the mean-field assumption.}
Real embedding distributions such as GloVe and SIFT exhibit substantial non-uniformity, with cluster structure, manifold geometry, and density variations corresponding to semantic structure. The assumption is therefore not literally satisfied by any of the datasets we evaluate. What the assumption requires \emph{in practice} is that the optimizer selects $(m, G)$ such that adjacent cells have similar local density on average, a weaker condition than global uniformity, and one that is implicitly enforced, since configurations straddling severe density gradients incur recall penalties and are filtered from the Pareto front. The assumption nevertheless degrades as $n_{\mathrm{probe}}$ grows, because probing further from the home cell increases the likelihood of crossing a density gradient. High recall therefore corresponds to the regime where mean-field is most strained, by the same mechanism that makes the Taylor truncation in~\eqref{eq:taylor_app} least accurate. The two limitations are not independent, both reflecting the algorithm operating outside the local neighborhood of the home cell (where the mean-field description applies). The predicted log-linear form nevertheless holds empirically over most of the operating range in Figure \ref{fig:figure1}, which we discuss further in Appendix~\ref{app:validity}.

\paragraph{Dropping $\ln \Delta$.} Under~\eqref{eq:meanfield_regime}, we drop $\ln \Delta$ from the argument of the outer log in~\eqref{eq:log_qps_taylor}. To confirm this is a well-controlled approximation: because $\ln \Delta < 0$,
\[
\frac{R}{R_{\max}} + \ln\Delta \;<\; \frac{R}{R_{\max}},
\]
and monotonicity of $\log$ gives
\[
\log\!\Big(R/R_{\max} + \ln\Delta\Big) \;<\; \log(R/R_{\max}),
\]
so $\log(R/R_{\max})$ is a valid upper bound, with the gap shrinking to zero as $\mu\theta \to 0$. Applying this approximation to~\eqref{eq:log_qps_taylor},
\begin{equation}
\label{eq:log_qps_no_delta}
\log \mathrm{QPS}(R) \;\approx\; \log(\mathrm{const}) \;-\; \log(R/R_{\max}).
\end{equation}

\paragraph{Expanding the log-ratio.} Expanding $\log(R/R_{\max}) = \log R - \log R_{\max}$ and substituting into~\eqref{eq:log_qps_no_delta},
\begin{equation}
\log \mathrm{QPS}(R) \;\approx\; \log(\mathrm{const}) - \log R + \log R_{\max}
\;=\; \log(\mathrm{const} \cdot R_{\max}) - \log R.
\end{equation}
Absorbing $\log R_{\max}$ into the constant, we define $\mathrm{const}' := \mathrm{const} \cdot R_{\max}$ and obtain
\begin{equation}
\label{eq:log_qps_powerlaw}
\log \mathrm{QPS}(R) \;\approx\; \log(\mathrm{const}') - \log R.
\end{equation}
Equation~\eqref{eq:log_qps_powerlaw} is a power-law relationship, $\mathrm{QPS} \propto 1/R$.

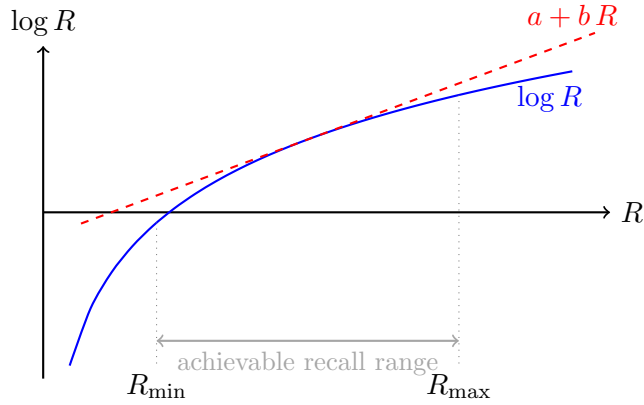
\begin{figure}[H]
\centering
\begin{tikzpicture}[scale=1.0]
  \draw[->, thick] (0,0) -- (7.5,0) node[right] {$R$};
  \draw[->, thick] (0,-2.2) -- (0,2.2) node[above] {$\log R$};

  \draw[thick, domain=0.35:7, smooth, variable=\x, blue]
      plot ({\x}, {1.3*ln(0.6*\x)});
  \node[blue] at (6.7, 1.5) {$\log R$};

  \draw[thick, dashed, red, domain=0.5:7.3, variable=\x]
      plot ({\x}, {0.964 + 0.371*(\x - 3.5)});
  \node[red] at (7.0, 2.6) {$a + b\,R$};

  \draw[dotted, gray] (1.5, -2.0) -- (1.5, {1.3*ln(0.6*1.5)});
  \draw[dotted, gray] (5.5, -2.0) -- (5.5, {1.3*ln(0.6*5.5)});
  \node[below] at (1.5, -2.0) {$R_{\min}$};
  \node[below] at (5.5, -2.0) {$R_{\max}$};

  \draw[<->, thick, gray!70]
      (1.5, -1.7) -- (5.5, -1.7);
  \node[gray!70, below] at (3.5, -1.7) {\small achievable recall range};

\end{tikzpicture}
\caption{Local linearization of $\log R$ on the achievable recall interval $[R_{\min}, R_{\max}]$. The concave $\log R$ curve (blue, solid) is approximated by a tangent line $a + b\,R$ (red, dashed) over the narrow operating range of the algorithm. Substituting this linear approximation into $\log \mathrm{QPS} \approx \log(\mathrm{const}\cdot R_{\max}) - \log R$ converts the power-law dependence into an exponential one, yielding $\mathrm{QPS}(R) \approx A' e^{-B' R}$.}
\label{fig:log_linearization}
\end{figure}

\paragraph{Local linearization of $\log R$.} To recover the exponential form observed empirically, we exploit the fact that $R$ is confined to a bounded interval $[R_{\min}, R_{\max}]$ and linearize:
\begin{equation}
\label{eq:logR_linear}
\log R \;\approx\; a + b\,R, \qquad R \in [R_{\min}, R_{\max}],
\end{equation}
with $b > 0$ because $\log$ is increasing. Geometrically, this replaces the concave $\log R$ curve with a tangent line over the achievable recall range, as shown in Figure~\ref{fig:log_linearization}. Substituting~\eqref{eq:logR_linear} into~\eqref{eq:log_qps_powerlaw},
\begin{equation}
\label{eq:log_qps_linear}
\log \mathrm{QPS}(R) \;\approx\; \log(\mathrm{const}') - a - b\,R.
\end{equation}

\paragraph{Exponentiating.} Since $\log = \log_{10}$ throughout, exponentiating~\eqref{eq:log_qps_linear} in base 10 yields
\begin{equation}
\mathrm{QPS}(R) \;\approx\; 10^{\log(\mathrm{const}') - a - b R} \;=\; \mathrm{const}' \cdot 10^{-a} \cdot 10^{-b R}.
\end{equation}
Absorbing the two $R$-independent factors into a single constant $A' := \mathrm{const}' \cdot 10^{-a}$ and setting $B := b$, we obtain the log-linear throughput--recall relation:
\begin{equation}
\label{eq:qps_final}
\boxed{\;\mathrm{QPS}(R) \;\approx\; A' \cdot 10^{-B R}, \qquad A' > 0,\ B > 0.\;}
\end{equation}
Equivalently, $\mathrm{QPS}(R) \approx A' e^{-B' R}$ with $B' = B \ln 10$, confirming the log-linear relationship observed empirically in Section~\ref{sec:pareto}. We note that the slope $B$ is set by the local derivative of $\log R$ on $[R_{\min}, R_{\max}]$, which scales inversely with $R_{\max}$.

\subsection{Validity, robustness, and empirical correspondence}
\label{app:validity}

The closed form~\eqref{eq:qps_final} relies on multiple approximations: a Taylor truncation~\eqref{eq:taylor_app}, a mean-field assumption~\eqref{eq:meanfield_regime} that real embedding distributions do not fully satisfy, and a local linearization of $\log R$ (\eqref{eq:logR_linear}). Taylor truncation and mean-field degrade in the high-recall regime, which is also the regime where the empirical $d$-scaling crossover reported in the main text emerges. Here, we argue that the qualitative log-linear form is structural and robust to its underlying assumptions, with Figure~\ref{fig:figure1} corroborating our reasoning. In addition, the empirical $d$-scaling characterization in the main text is methodologically independent of the closed-form model.

\paragraph{Robustness of the log-linear form.}
It is not obvious \emph{a priori} that~\eqref{eq:qps_final} should describe the empirical Pareto front at all, given that the mean-field assumption is not fully satisfied by the non-uniform embedding distributions on which we evaluate the algorithm. We argue that the qualitative log-linear shape is structural rather than approximation-dependent. Two mechanical features of the algorithm produce the log-linear shape: (i) recall is a saturating, monotonically increasing function of $n_{\mathrm{probe}}$, because cells are probed in order of decreasing membership probability $P_i$; and (ii) cost is linear in $n_{\mathrm{probe}}$, because each probed cell contributes a fixed expected candidate count $N/G^m$. Combining a saturating-concave recall function with a linear cost function yields cost growing faster than linearly in recall as recall approaches its ceiling, which on log-y vs.\ linear-recall axes traces out a near-straight line over the operating range with downward curvature near the ceiling. Importantly, neither (i) nor (ii) requires the mean-field assumption. (i) requires only that cells can be ordered by membership probability, which the algorithm enforces by construction, and (ii) follows from each probed cell contributing a fixed expected candidate count. Thus, the approximations affect the predicted slope $B$ and ceiling $R_{\max}$, but not the functional form itself, which we observe on real, non-uniform embedding data.

\paragraph{Relevance to the $d$-scaling crossover.}
The closed form and the $d$-scaling characterization reported in Figure~\ref{fig:figure2}b answer fundamentally different questions. While ~\eqref{eq:qps_final} models the shape of the QPS--recall Pareto front for a fixed dataset, $\alpha_d$ characterizes how QPS (at fixed recall) scales across datasets of different dimensionality. The closed form contains no $d$-dependence in its functional form. Rather, $d$ manifests empirically by influencing the Pareto-optimal $(m, G, n_{\mathrm{probe}})$ configuration. Thus, the absence of explicit $d$-dependence in~\eqref{eq:qps_final} underscores that the $d$-scaling crossover is an authentic, empirical finding that reflects the algorithm's ability to adapt to dataset dimensionality. 

\newpage

\section{Implementation and Experimental Details} 
\label{sec:methods}
\renewcommand{\thetable}{C\arabic{table}}
\setcounter{table}{0}

\subsection{Multiprobe grid search}

\subsubsection{Index construction}
We employ PCA to project each point in $\sD$ from $\R^{d}$ to $\R^{m}$, where the grid dimensionality $m \ll d$ is a hyperparameter. The projected space is partitioned into a uniform grid of $G^m$ cells, where $G$ is the number of splits per dimension. Each cell stores the indices of points whose projections fall within its boundaries. To handle empty cells encountered during querying, we precompute a multi-source breadth-first search (BFS) over the grid graph in which every occupied cell is seeded simultaneously as a source. The BFS expands outward through cell neighbors, recording for each cell the identity of its nearest occupied cell. This process is easy to parallelize and each non-empty cell only needs to be updated once. The result is a static lookup table mapping every grid cell to the nearest occupied cell by Manhattan distance. Insertion and deletion of new points are also efficient under this regime. Figure \ref{fig:figureD12} shows examples of how often the BFS fallback is triggered as a function of dataset characteristics (i.e., $N$ and $d$) and hyperparameters ($G$ and $n_{\text{probe}}$). 

\subsubsection{Query processing}
Given a query vector $\vq \in \R^d$ and a probe count $n_\text{probe} \geq 1$, we project $\vq$ to the $m$-dimensional PCA space and identify the cell containing the projected vector $\vq'$. When $n_\text{probe} = 1$, only the primary cell is searched. When $n_\text{probe} > 1$, we additionally probe the $n_\text{probe} - 1$ nearest neighboring cells ranked by wall distance (the squared Euclidean distance from $\vq'$ to the shared boundary of each neighboring cell). To bound the search space, we consider only the $2^m$ cells in the geometric orthant (the $m$-dimensional generalization of a quadrant) toward which $\vq'$ is displaced from the primary cell's midpoint. Candidates from all probed cells are gathered and ranked using the full $d$-dimensional metric (cosine similarity for angular datasets, Euclidean distance for $L_2$ datasets) to return the top-$k$ results. If all probed cells are empty, the precomputed BFS fallback is triggered to return the nearest occupied cell.

\subsection{Experimental setup}
All experiments use the \texttt{ann-benchmarks} framework \cite{aumuller_ann-benchmarks_2020, aumuller_reproducibility_2021}, which provides standardized evaluation with Docker containers enforcing single-CPU execution per algorithm. We report recall@$k=10$ as the accuracy metric and queries per second (QPS) as the throughput metric.

\subsubsection{Baselines}  
We compare the multiprobe grid search with the following algorithms: Voyager (a HNSW variant, graph-based) \cite{noauthor_spotifyvoyager_2026} , PyNNDescent (graph-based) \cite{mcinnes_lmcinnespynndescent_2026}, Annoy (tree-based) \cite{noauthor_spotifyannoy_2026}, and FAISS-IVF (quantization-based partitioning) \cite{noauthor_facebookresearchfaiss_2026}. We selected these state-of-the-art algorithms to represent the major ANN algorithm families. Brute-force search serves as a reference for several performance metrics (e.g., queries per second) and to identify ground truth nearest neighbors.

\subsubsection{Hyperparameter tuning for multiprobe grid} 
The multiprobe grid algorithm has three hyperparameters: grid dimensionality $m$, grid splits $G$, and probe count $n_\text{probe}$. The pair $(m, G)$ determines the index structure, while $n_\text{probe}$ impacts the recall-latency tradeoff at query time. We run NSGA-II multi-objective optimization (200 trials) using the Optuna package to identify Pareto-optimal $(m, G)$ pairs for the GloVe-200-angular dataset; the same configurations are applied to subsampled GloVe-200 and to GloVe-25, -50, and -100-angular. A separate optimization was performed on SIFT-128-euclidean because of the distinct data modality. From each Optuna study, representative $(m, G)$ pairs spanning recall $>$0.10 to 0.98 are selected for all subsequent experiments. For each pair, $n_{\text{probe}}$ is swept across a range of values up to $2^m$ to trace the recall-QPS frontier. Baseline algorithms use their respective \texttt{ann-benchmarks} default parameter sweeps.

\subsubsection{Scaling analysis} 
\label{scaling_analysis}
We characterize scaling behavior by fitting $\log(\text{QPS}) = \alpha_x \log(x) + b$, where $x$ corresponds to either dataset size ($N$) or dataset dimensionality ($d$). The exponent $\alpha_x$ captures how query throughput changes with $x$. For each algorithm, dataset, and recall target, we construct  the Pareto front over all benchmarked configurations and   interpolate QPS to the target recall by linear interpolation in $\log_{10}(\mathrm{QPS})$ between the two Pareto-optimal configurations whose measured recalls bracket the target. If the target recall falls outside the Pareto front's recall    range for a given dataset, that point is omitted from the fit.

\begin{itemize}
  \item For $N$-scaling studies, GloVe-200-angular is sampled at $N \in \{10^4, 2.0\times10^4, 4.0\times10^4, 7.5\times10^4, 1.5\times10^5, 3.0\times10^5, 6.0\times10^5, 1.18\times10^6\}$ points. SIFT-128-euclidean is sampled at $N \in \{10^4, 5\times10^4, 10^5, 5\times10^5, 10^6\}$ points. Ground truth ($k=10$ nearest neighbors) is recomputed for each subsample via brute-force.   
  \item $d$-scaling uses the GloVe word embedding family \cite{pennington_glove_2014} (GloVe-25, 50, 100, 200), which
provides four datasets of identical size ($N=1.18\times10^6$) and source distribution with varying native dimensionality $d \in \{25, 50, 100, 200\}$.
\end{itemize}

\subsection{Profiling analysis of multiprobe grid}

Our multiprobe grid implementation is in Python, while baseline algorithms (Voyager, PyNNDescent, Annoy, FAISS-IVF) use highly optimized C++. Here, we investigate whether Python interpreter overhead, memory allocation, and orchestration loops have significant impact on our reported scaling exponents.

\paragraph{First-principles reasoning.} The per-query cost primarily decomposes into two components: an algorithmic component involving calls into NumPy/BLAS C kernels (e.g., PCA projection, candidate gathering via concatenation and fancy indexing, re-ranking), and an overhead component involving interpreter-bound orchestration (e.g., primary cell lookup, neighbor cell enumeration, candidate-list assembly). The interpreter-bound operations operate on a fixed number of cells (at most $2^m$) and are therefore independent of $N$, contributing a constant, additive term to the per-query cost. In the regime where the algorithmic term dominates, which we expect for any modern dataset of appreciable $N$, this additive overhead becomes negligible and does not affect the slope of a log--log fit. The reported scaling exponents therefore reflect algorithmic complexity, not implementation language. 
  
\paragraph{Empirical profiling.} We profiled the multiprobe query path with \texttt{cProfile} on GloVe-200-angular across three $N$ scales spanning two orders of magnitude, at a representative configuration ($m{=}6$, $G{=}5$, $n_{\text{probe}}{=}16$, 1{,}000 queries per $N$). The query was decomposed into named phase functions so \texttt{cProfile}'s per-function attribution gives per-phase timings (Table~\ref{tab:profile}). Phases dominated by NumPy/BLAS calls (PCA projection, candidate gathering via \texttt{np.concatenate} and fancy indexing, and re-ranking via norm/matmul/argpartition) scale near-linearly with candidate count. In contrast, phases that are pure Python orchestration (cell index lookup, neighbor enumeration, list assembly) remain at $O(1 \space \mu s)$ cost across the entire range of $N$. Altogether, we observe a factor of $\sim 2.6$ increase from $N{=}10^4$ to $N{=}1.18 \times 10^6$ in the Python overhead and a factor of $\sim 865\times$ growth of the NumPy/BLAS-backed phases. At $N{=}1.18 \times 10^6$, Python orchestration accounts for under 0.01\% of total query time. The reported scaling exponents in the main text should therefore hold whether multiprobe grid is implemented in Python or C++.
          
\begin{table}[h]                                                                                                                                                                                                                                       
\centering      
\caption{Per-query phase times (\textmu s) for multiprobe grid on GloVe-200-angular ($m{=}6$, $G{=}5$, $n_{\text{probe}}{=}16$), profiled with \texttt{cProfile} over 1{,}000 queries per $N$.}                                                                                                                                                     
\label{tab:profile}
\small                                                                                                                                                                                                                                                             
\begin{tabular}{llccc}
\toprule
Phase & Bucket & $N{=}10^4$ & $N{=}1.5{\times}10^5$ & $N{=}1.18{\times}10^6$ \\                                                                                                                                                                                    
\midrule                                                                                                                                                                                                                                                           
PCA projection & NumPy/BLAS & 1.4 & 4.8 & 8.9 \\                                                                                                                                                                                                                         
Cell index lookup & Python & 0.3 & 0.8 & 1.1 \\                                                                                                                                                                                                                    
Neighbor enumeration & Python & 0.9 & 1.5 & 2.3 \\                                                                                                                                                                                                                 
Assemble candidate indices & Python & 0.2 & 0.4 & 0.6 \\                                                                                                                                                                                                                      
Gather candidate vectors & NumPy/BLAS & 64.9 & 5{,}354.3 & 62{,}334.9 \\                                                                                                                                                                                                     
Re-rank (norm/matmul/argpartition) & NumPy/BLAS & 9.4 & 311.9 & 3{,}125.7 \\                                                                                                                                                                                             
\midrule                                                                                                                                                                                                                                                           
\textbf{NumPy/BLAS subtotal} & & \textbf{75.7} & \textbf{5{,}670.9} & \textbf{65{,}469.6} \\                                                                                                                                                                             
\textbf{Python orchestration subtotal} & & \textbf{1.5} & \textbf{2.7} & \textbf{3.9} \\                                                                                                                                                                           
Total (sum of phases) & & 77.2 & 5{,}673.6 & 65{,}473.5 \\                                                                                                                                                                                                
\bottomrule                                                                                                                                                                                                                                                        
\end{tabular}                                                                                                                                                                                                                                                      
\end{table}

\subsection{Derivation of the rebuild-to-query crossover}                                                                                                                                                                                                               
\label{sec:crossover}
          
Section~\ref{sec:build} reports a rebuild-to-query crossover threshold for GloVe-200 of approximately one rebuild per 2,600--20,400 queries depending on the baseline. Here, we derive that range from the cost-framework from Sun et al \cite{sun_scaling_2025}.                        

\paragraph{Crossover condition.} The total cost is                                                                                                                                                                                                                       
\begin{equation}                                             
J(N) = f_I \cdot I(N) + f_C \cdot C(N) + M(N),
\end{equation}                                                                                                                                                                                                                                                           
where $f_I$ and $f_C$ denote the frequencies of index rebuild and query, respectively. Multiprobe grid achieves lower total cost than a given baseline when $J_{\text{grid}} < J_{\text{baseline}}$. Because the memory term $M(N)$ enters identically on both sides, it
cancels and the inequality reduces to                                                                                                                                                                                                                                    
\begin{equation}                                             
f_I \cdot I_{\text{grid}} + f_C \cdot C_{\text{grid}} \;<\; f_I \cdot I_{\text{baseline}} + f_C \cdot C_{\text{baseline}}.                                                                                                                                               
\end{equation}                                                                                                                                                                                                                                                           
Collecting the $f_I$ terms on the right and the $f_C$ terms on the left,
\begin{equation}                                                                                                                                                                                                                                                         
f_C \cdot \bigl(C_{\text{grid}} - C_{\text{baseline}}\bigr) \;<\; f_I \cdot \bigl(I_{\text{baseline}} - I_{\text{grid}}\bigr).
\end{equation}                                                                                                                                                                                                                                                           
Multiprobe has lower indexing cost ($I_{\text{grid}} < I_{\text{baseline}}$) and higher per-query latency ($C_{\text{grid}} > C_{\text{baseline}}$), so we define $\Delta I := I_{\text{baseline}} - I_{\text{grid}} > 0$ and $\Delta C := C_{\text{grid}} -             
C_{\text{baseline}} > 0$. Both sides of the inequality are positive. Dividing both sides by $f_I \cdot \Delta C > 0$ (which preserves the inequality direction),                                                                                                         
\begin{equation}                                                                                                                                                                                                                                                         
\frac{f_C}{f_I} \;<\; \frac{\Delta I}{\Delta C} \;=:\; N_{\text{cross}}.                                                                                                                                                                                                 
\end{equation}                                                                                                                                                                                                                                                           
The ratio $f_C/f_I$ is the expected number of queries per index rebuild; multiprobe wins on total cost whenever this ratio falls below the crossover $N_{\text{cross}}$.
          
\paragraph{Empirical crossovers for GloVe-200 ($N=1.18\times10^6$).} For each algorithm we use the Pareto-optimal configuration nearest to recall@$k$=10 ${=}0.80$. Multiprobe's anchor is $(m{=}6, G{=}5)$. Per-algorithm values and the resulting crossovers are reported in        
Table~\ref{tab:crossover}; $\Delta I$ and $\Delta C$ are converted to a common time unit (seconds) before computing the ratio. The 2,600--20,400 range cited in Section~\ref{sec:build} is set by the FAISS-IVF crossover (2,591) and the Voyager crossover (20,432), the
two extremes among the four baselines.                                                                                                                                                                                                                                  
                                                           
\begin{table}[h]
\centering
\caption{Rebuild-to-query crossover threshold $N_{\text{cross}}$ for multiprobe grid against each baseline at $N=1.18\times10^6$, recall@$k$=10 ${=}0.80$. Multiprobe achieves lower total cost when $f_C/f_I < N_{\text{cross}}$ (i.e., when index rebuilds happen at
least once per $N_{\text{cross}}$ queries).}                                                                                                                                        
\label{tab:crossover}
\small                                                                                                                                                                                                                                                                   
\begin{tabular}{lccccr}                                                                                                                                                                                                                                                  
\toprule
Algorithm & $I$ (s) & $C$ (ms) & $\Delta I$ (s) & $\Delta C$ (ms) & $N_{\text{cross}}$ \\                                                                                                                                                                                
\midrule                                                                                                                                                                                                                                                                 
\textbf{Multiprobe Grid} & \textbf{8.4} & \textbf{76.98} & --- & --- & --- \\
\midrule                                                                                                                                                                                                                                                                 
FAISS-IVF & 206 & 0.72 & 197.6 & 76.26 & 2{,}591 \\          
Annoy & 333 & 7.20 & 324.6 & 69.78 & 4{,}652 \\                                                                                                                                                                                                                          
PyNNDescent & 500 & 0.56 & 491.6 & 76.42 & 6{,}431 \\                                                                                                                                                                                                           
Voyager (HNSW) & 1{,}569 & 0.62 & 1560.6 & 76.36 & 20{,}432 \\          
 
\bottomrule                                                                                                                                                                                                                                                              
\end{tabular}                                                                                                                                                                                                                                                            
\end{table}              

\newpage

\section{Supplementary Figures} 
\renewcommand{\thefigure}{D\arabic{figure}}
\setcounter{figure}{0}

\begin{figure}[h]
    \centering
    \includegraphics[width=1.0\linewidth]{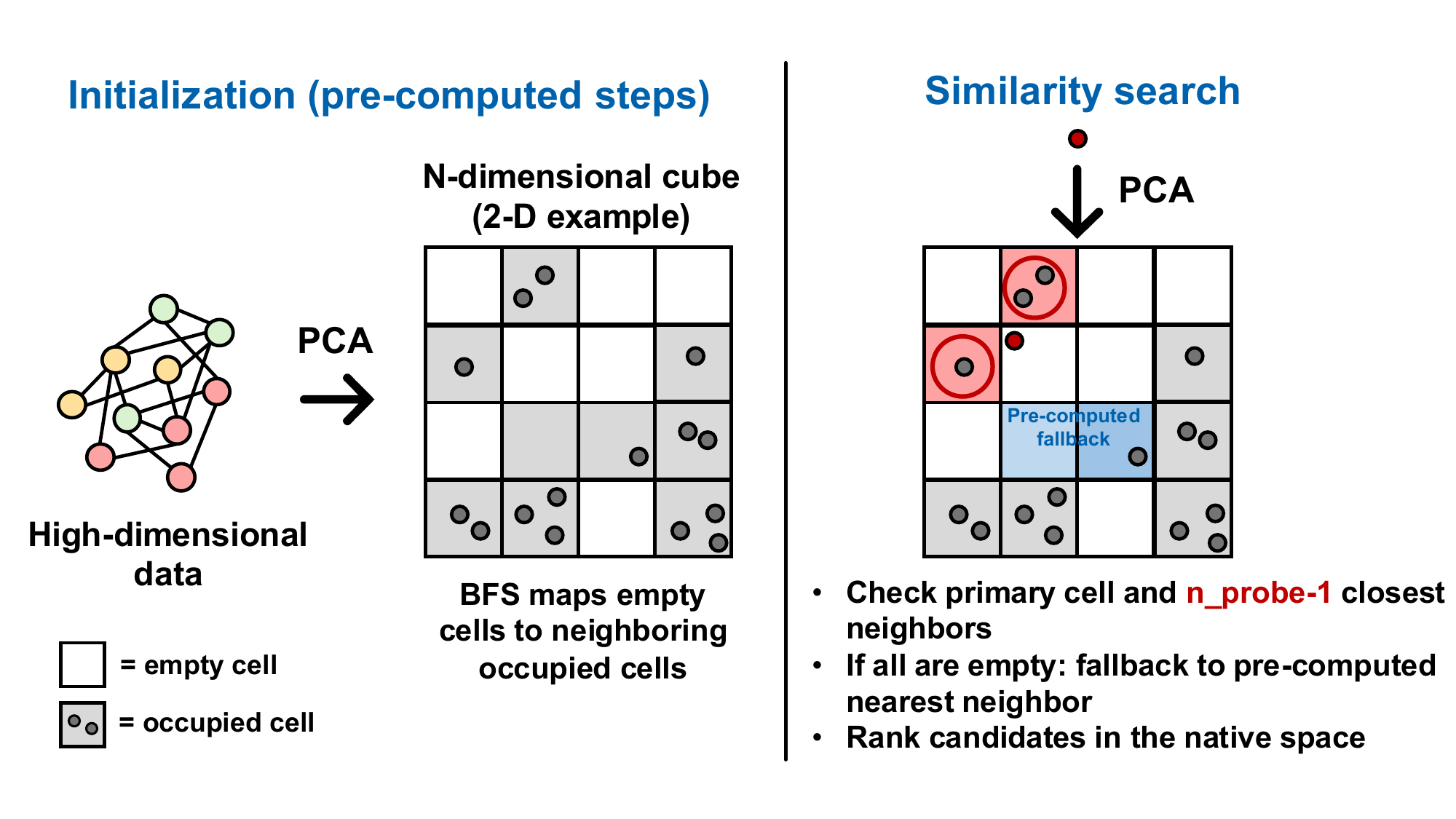} 
    \caption{\underline{Overview of the multiprobe grid algorithm.} \textbf{Initialization:} dataset points are projected from $\mathbb{R}^d$ to $\mathbb{R}^m$ ($m \ll d$) via principal component analysis and assigned to cells in a uniform $G^m$ grid. Empty cells store a pointer to the nearest occupied cell, as computed by multisource breadth-first-search. \textbf{Similarity search:} the query $\mathbf{q}$ is projected to $\mathbf{q}'$, which identifies a primary cell. For $n_{\text{probe}} > 1$, additional cells in the orthant toward which $\mathbf{q}'$ is displaced are probed in order of wall distance. Candidates from all probed cells are gathered and re-ranked in the native $d$-dimensional space to return the top-$k$ neighbors.}
    \label{fig:figureD1} 
\end{figure}
\clearpage

\begin{figure}[h]
    \centering
    \includegraphics[width=0.70\linewidth]{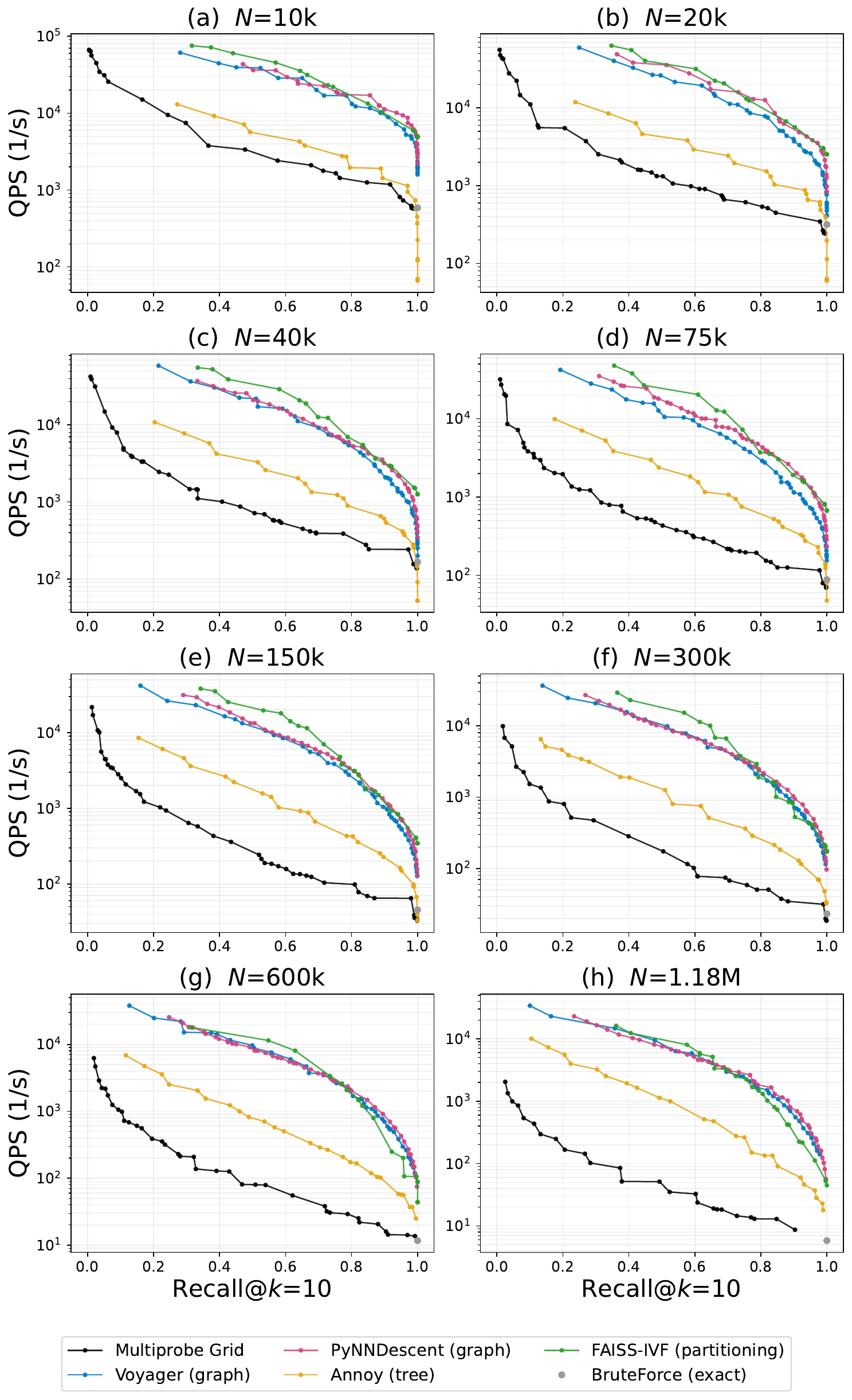}
    \caption{Pareto fronts across subsampled $N$ on GloVe-200-angular ($d=200$), where each panel shows all five algorithms + brute-force (which appears as a single point at recall=1.0). Note that panel (h) corresponds to Figure \ref{fig:figure1} in the main text.}
    \label{fig:figureD2}
\end{figure}
\clearpage

\begin{figure}[h]
    \centering
    \includegraphics[width=1.0\linewidth]{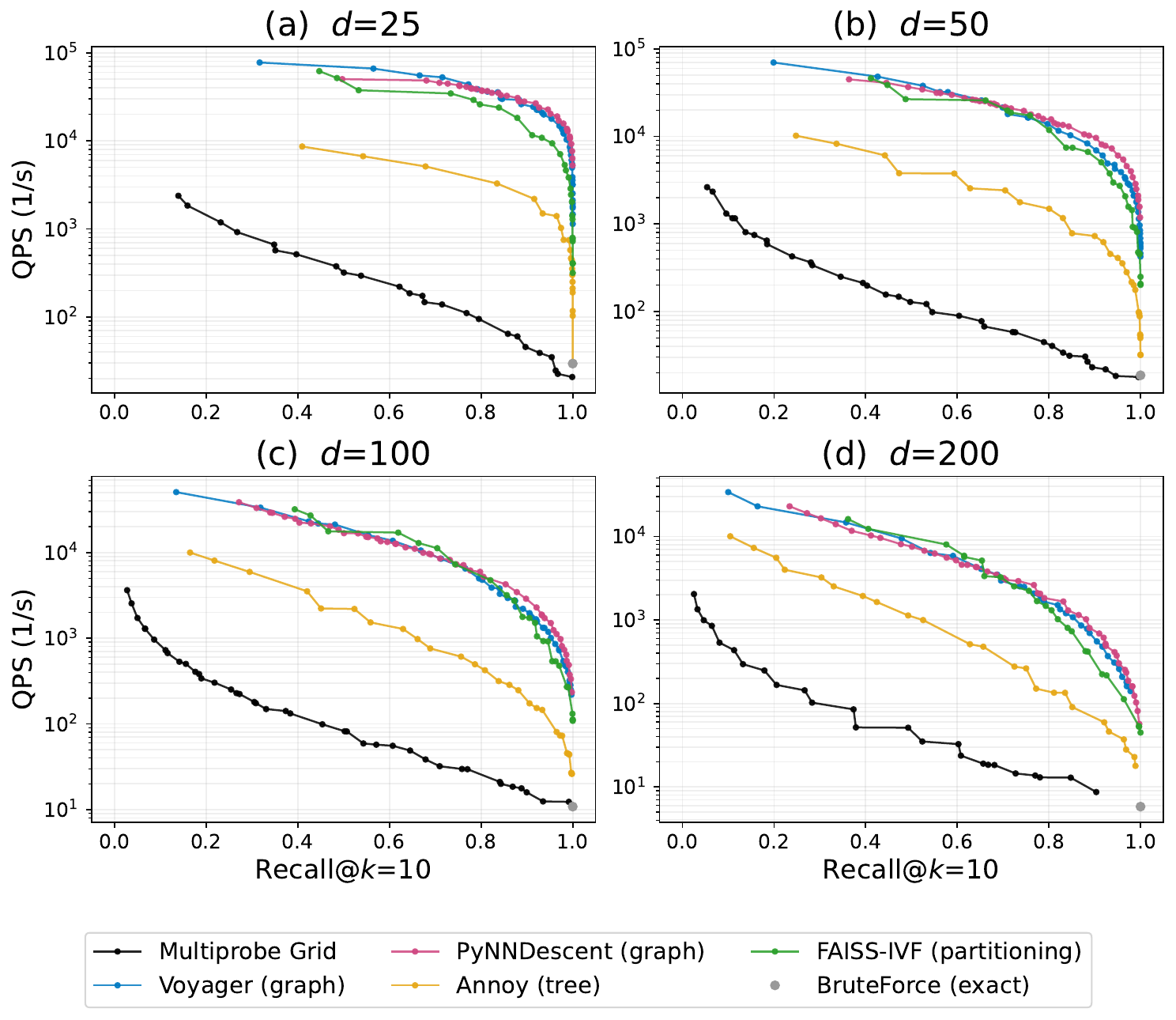}
    \caption{Pareto fronts across the GloVe-25-, 50-, 100-, and 200-angular datasets ($N=1.18\times10^6$), where each panel shows all five algorithms + brute-force (which appears as a single point at recall=1.0). Note that panel (d) corresponds to panel (h) in Figure \ref{fig:figureD2} and Figure \ref{fig:figure1} in the main text.}  
    \label{fig:figureD3}
\end{figure}
\clearpage

\begin{figure}[h]
    \centering
    \includegraphics[width=0.95\linewidth]{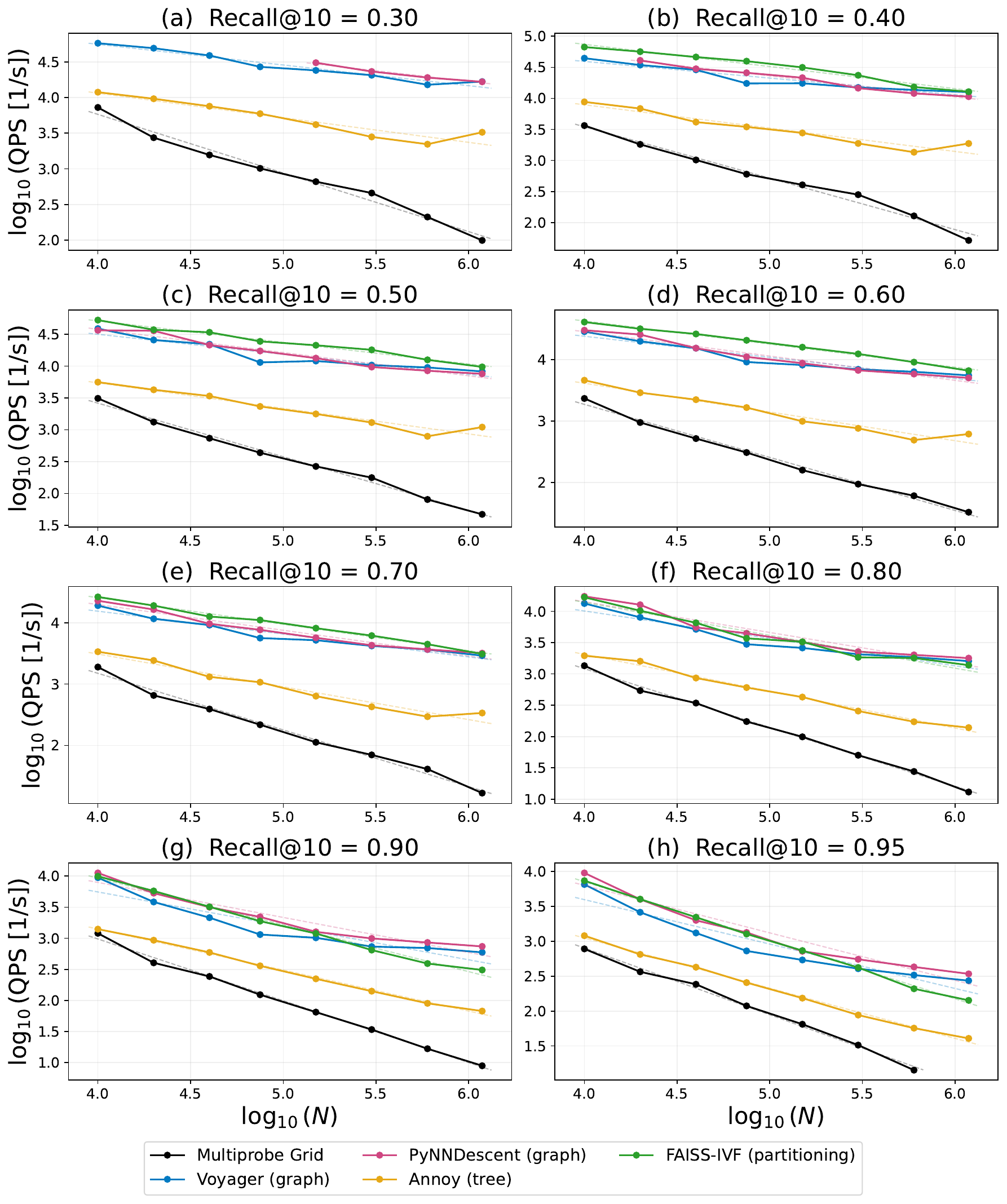}
    \caption{$\log_{10}(\mathrm{QPS})$ vs. $\log(N)$ plots at each target recall@$k$=10 used to derive the exponents in Figure \ref{fig:figure2}a. Datasets: GloVe-200-angular ($d=200)$ subsampled at $N=10^4$, $2.0\times10^4$, $4.0\times10^4$, $7.5\times10^4$, $1.5\times10^5$, $3.0\times10^5$, $6.0\times10^5$, and the full $1.18\times10^6$ points. The slope of each curve gives an $\alpha_N$ value for the respective algorithm and target recall (one data point in Figure~\ref{fig:figure2}a). Note that QPS values are interpolated along the Pareto front for each $N$ and target recall. Points are omitted where no configuration in the benchmark sweep achieved the target recall.}
    \label{fig:figureD4}
\end{figure}
\clearpage

\begin{figure}[h]
  \centering
  \includegraphics[width=1.0\linewidth]{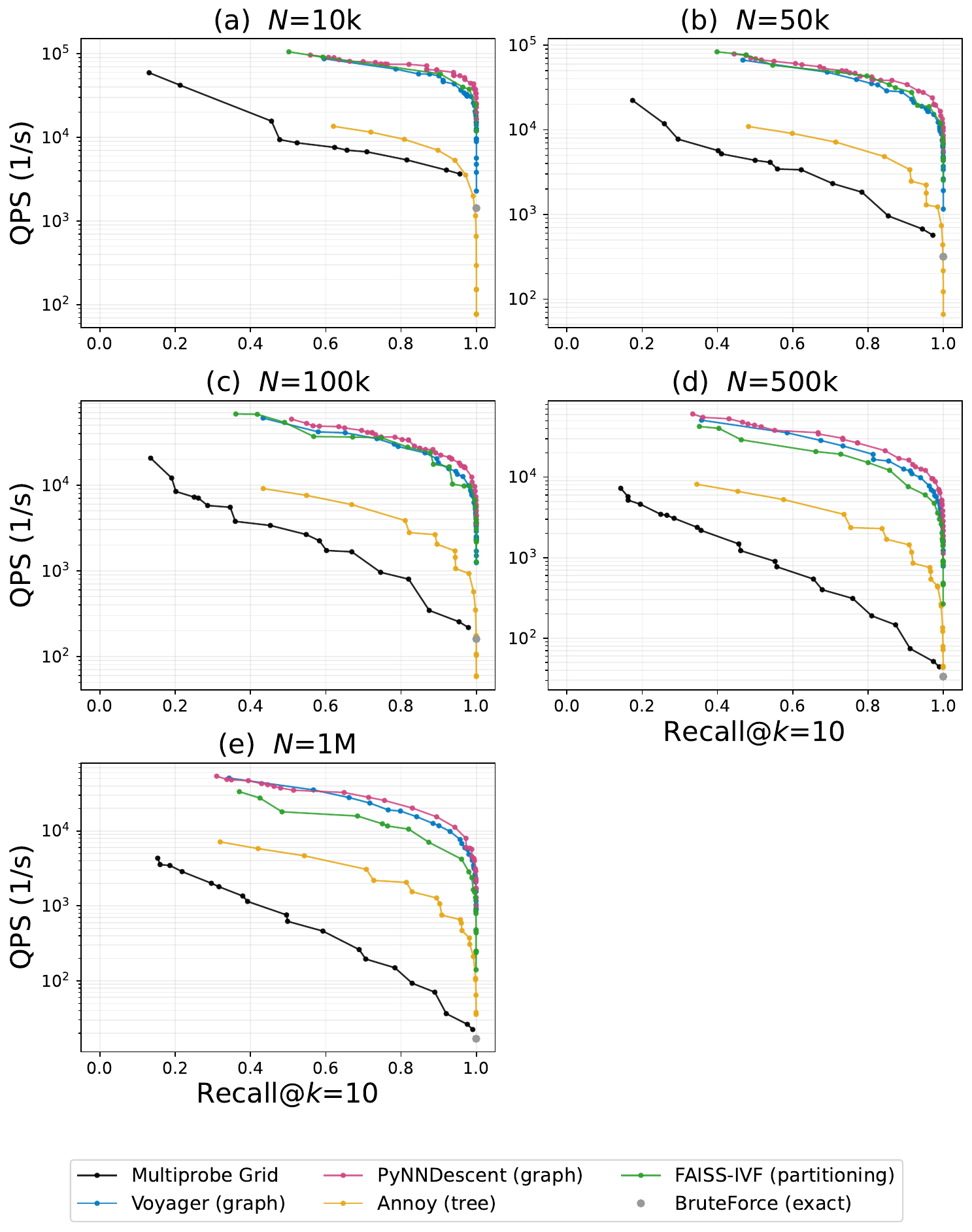}
  \caption{Pareto fronts across subsampled $N$ on SIFT-128-euclidean ($d=128$), where each panel shows all five algorithms + brute-force (which appears as a single point at recall=1.0).}
  \label{fig:figureD5}                                       
\end{figure}                                           
\clearpage

\begin{figure}[h]                                    
  \centering                                            
  \includegraphics[width=0.85\linewidth]{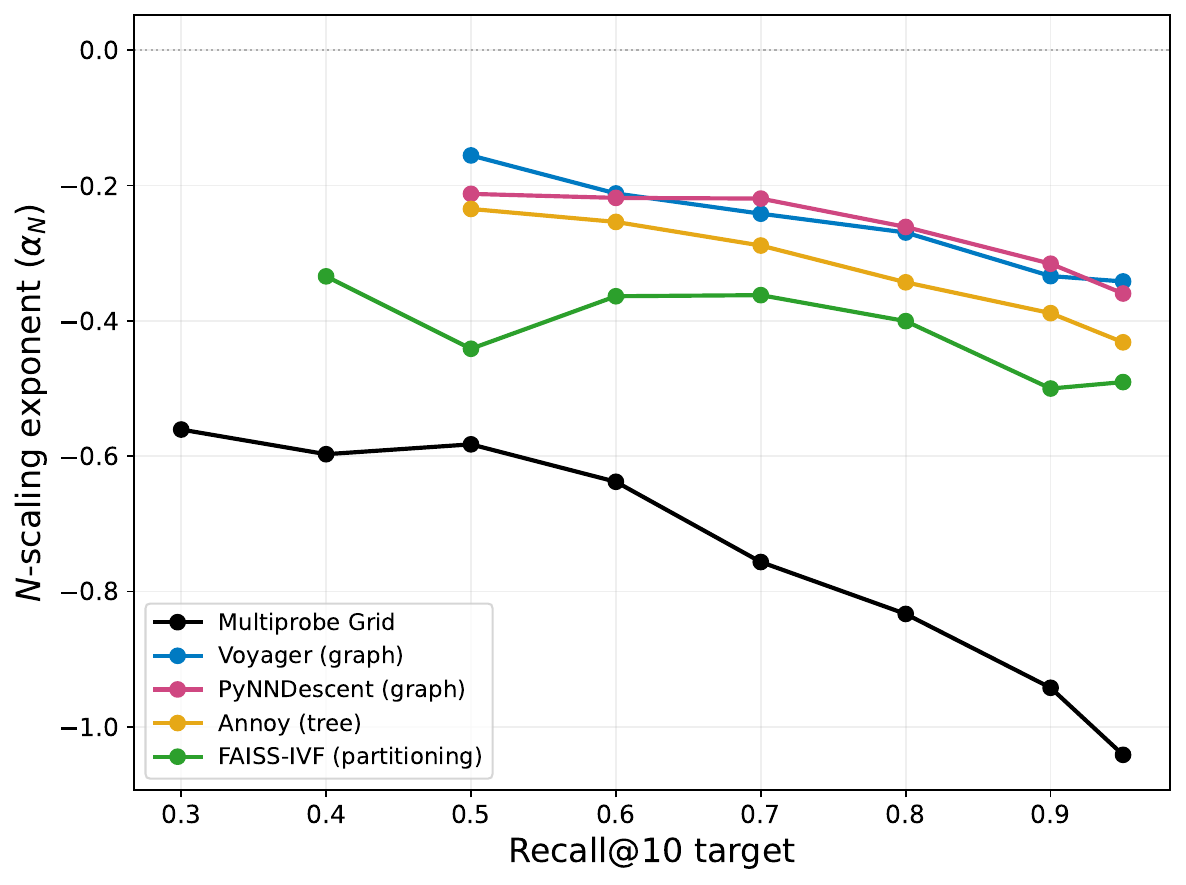}     
  \caption{$N$-scaling exponent $\alpha_N$ vs.\ recall@$k$=10 target on SIFT-128-euclidean. Datasets: SIFT-128-euclidean subsampled at $N \in \{10^4, 5\times10^4, 10^5, 5\times10^5, 10^6\}$.}  
  \label{fig:figureD6}            
\end{figure}                                                                                                         \clearpage                                                                                                                    
                      
\begin{figure}[h] 
  \centering                   
  \includegraphics[width=0.95\linewidth]{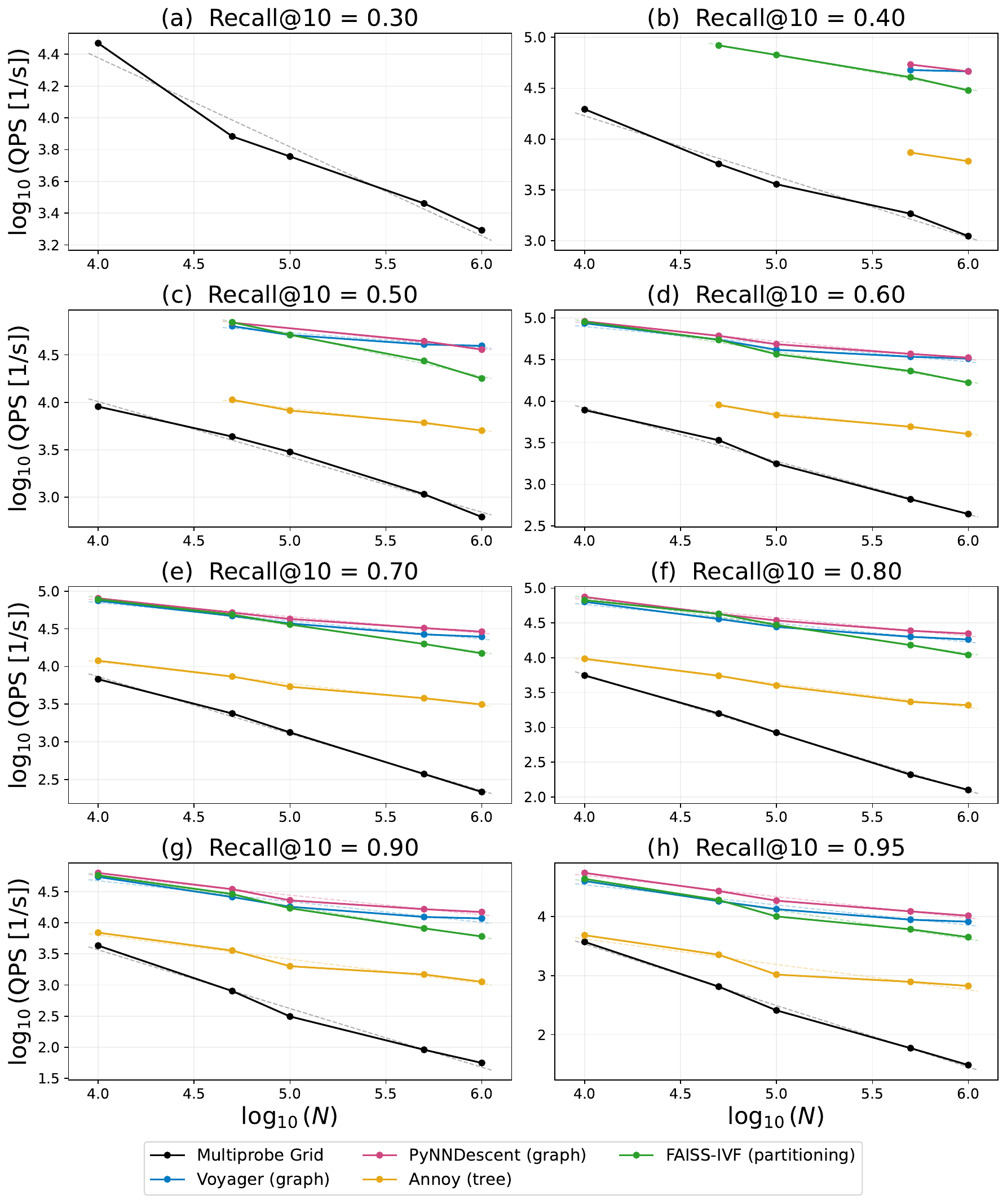}                      
  \caption{$\log_{10}(\mathrm{QPS})$ vs.\ $\log_{10}(N)$ plots at each target recall@$k$=10. Datasets: SIFT-128-euclidean ($d=128$) subsampled at $N \in \{10^4, 5\times10^4, 10^5, 5\times10^5, 10^6\}$. The slope of each curve gives an $\alpha_N$ value for the respective algorithm and target recall (one data point in Figure~\ref{fig:figureD6}). Note that QPS values are interpolated along the Pareto front for each $N$ and target recall. Points are omitted where no configuration in the benchmark sweep achieved the target recall}
  \label{fig:figureD7} 
\end{figure}   
\clearpage

\begin{figure}[h]
    \centering
    \includegraphics[width=0.95\linewidth]{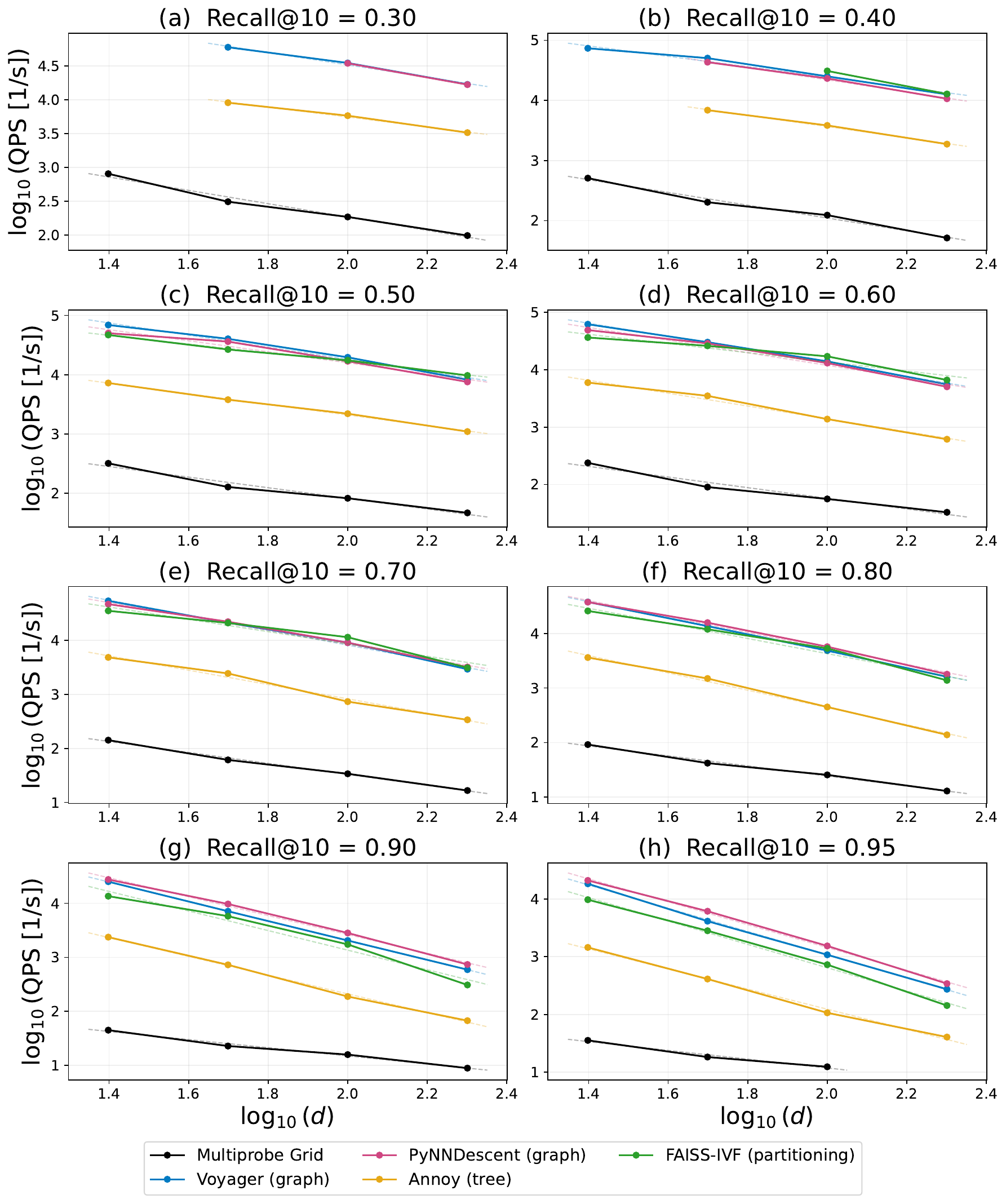}
    \caption{$\log_{10}(\mathrm{QPS})$ vs. $\log(d)$ plots at each target recall@$k$=10 used to derive the exponents in Figure \ref{fig:figure2}b. Datasets: GloVe-25, -50, -100, and -200-angular ($N=1.18\times10^6$). The slope of each curve gives an $\alpha_d$ value for the respective algorithm and target recall (one data point in Figure~\ref{fig:figure2}b). Note that QPS values are interpolated along the Pareto front for each $d$ and target recall. Points are omitted where no configuration in the benchmark sweep achieved the target recall.}
    \label{fig:figureD8}
\end{figure}
\clearpage

\begin{figure}[h]
    \centering
    \includegraphics[width=0.95\linewidth]{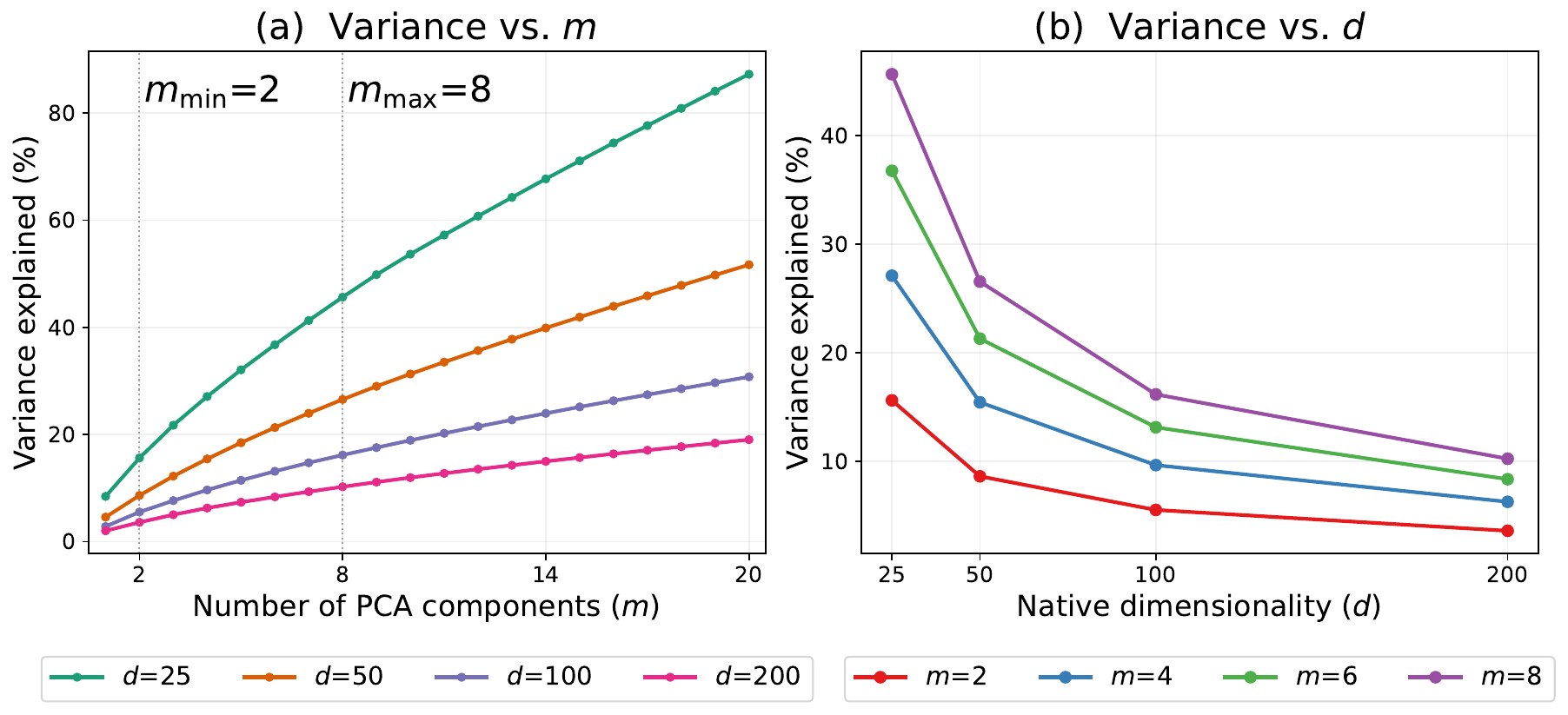}
    \caption{Variance retained by PCA projection as a function of the number of PCA components $m$ and native dimensionality of the dataset $d$. \textbf{(a)} Cumulative variance vs. $m$ for the GloVe-25, -50, -100, and -200-angular datasets ($N=1.18\times10^6$). Dotted vertical lines denote the minimum and maximum PCA components explored during hyperparameter optimization of multiprobe grid. \textbf{(b)} Variance explained at fixed $m \in \{2, 4, 6, 8\}$ as a function of native dimensionality $d$. Although the absolute variance explained with $m$ components decreases with $d$, the grid algorithm compensates by adapting its projection depth at query time (Table~\ref{tab:tableE1}).}
    \label{fig:figureD9}
\end{figure}
\clearpage

\begin{figure}[h]
    \centering
    \includegraphics[width=1.0\linewidth]{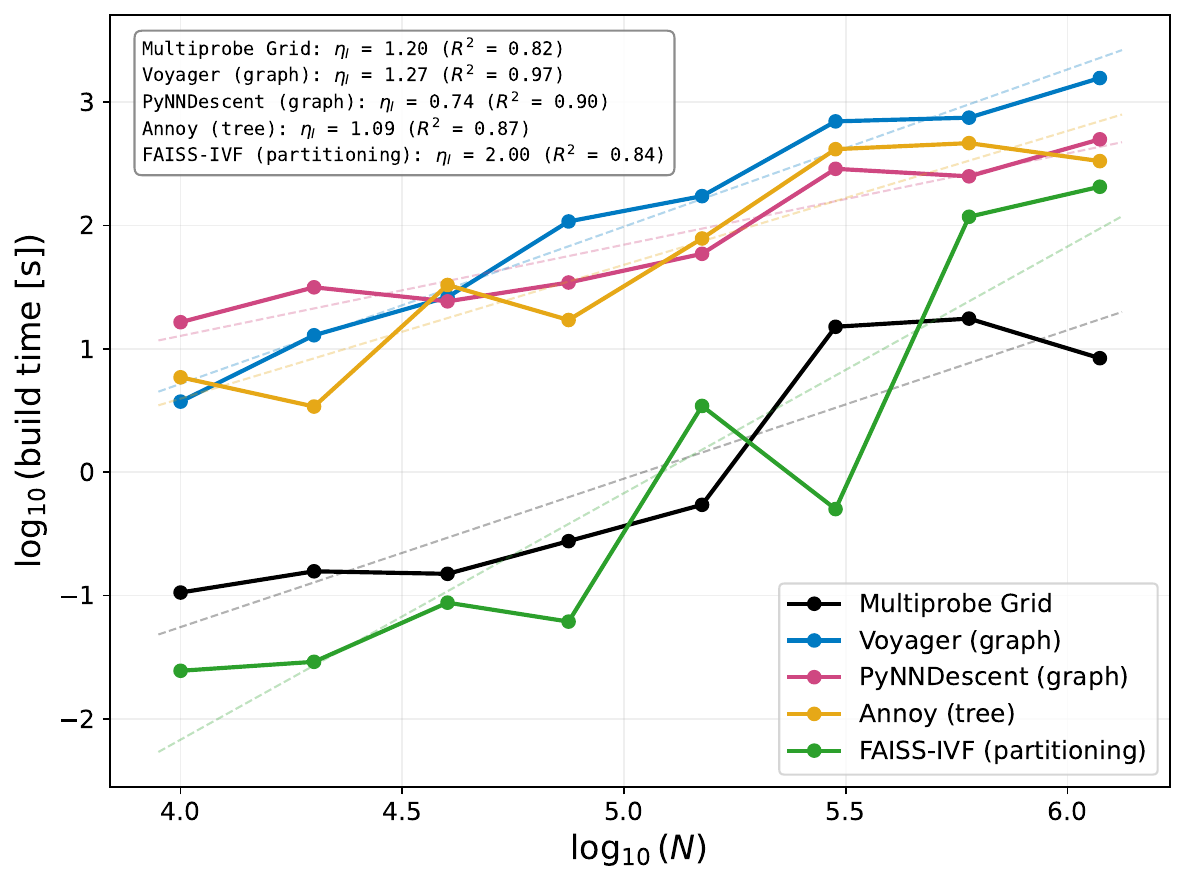}
    \caption{Build time scaling with $N$ across all five algorithms. Datasets: GloVe-200-angular ($d=200)$ subsampled at $N=10^4$, $2.0\times10^4$, $4.0\times10^4$, $7.5\times10^4$, $1.5\times10^5$, $3.0\times10^5$, $6.0\times10^5$, and the full $1.18\times10^6$ points. For each algorithm and $N$, we report the build time of the Pareto-optimal configuration nearest to recall@$k$=10 $= 0.80$; because this configuration can vary across $N$, fit quality (reported as $R^2$) differs across algorithms. The slope of each fit gives the indexing scaling exponent $\eta_I$, as reported in Table \ref{tab:table1} (the exponent for $I(N)$ entries).}
    \label{fig:figureD10}
\end{figure}
\clearpage

\begin{figure}[h]
    \centering
    \includegraphics[width=1.0\linewidth]{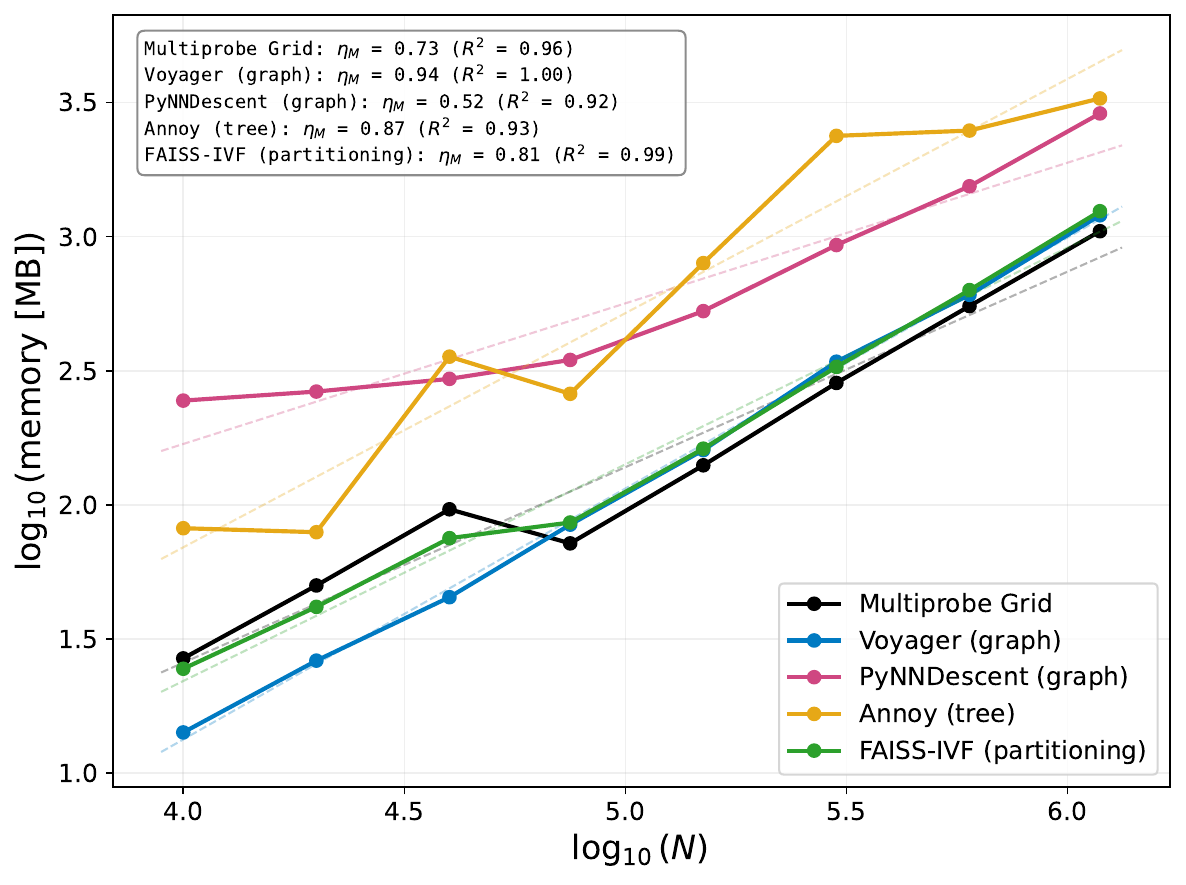}
    \caption{Memory footprint scaling with $N$ across all five algorithms. Datasets: GloVe-200-angular ($d=200)$ subsampled at $N=10^4$, $2.0\times10^4$, $4.0\times10^4$, $7.5\times10^4$, $1.5\times10^5$, $3.0\times10^5$, $6.0\times10^5$, and the full $1.18\times10^6$ points. Following \texttt{ann-benchmarks} \cite{aumuller_ann-benchmarks_2020}, memory footprint is measured as the difference in process resident set size (RSS) before and after index construction. We plot values for the Pareto-optimal configuration nearest to recall@$k$=10 $= 0.80$ at each $N$. The slope of each fit gives the memory scaling exponent $\eta_M$, as reported in Table \ref{tab:table1} (the exponent for $M(N)$ entries). Because memory footprint is measured via process RSS rather than on-disk index size, it may include differences in Python runtime and loaded data overhead in addition to the index itself.}
    \label{fig:figureD11}
\end{figure}
\clearpage

\begin{figure}[h]
    \centering
    \includegraphics[width=1.0\linewidth]{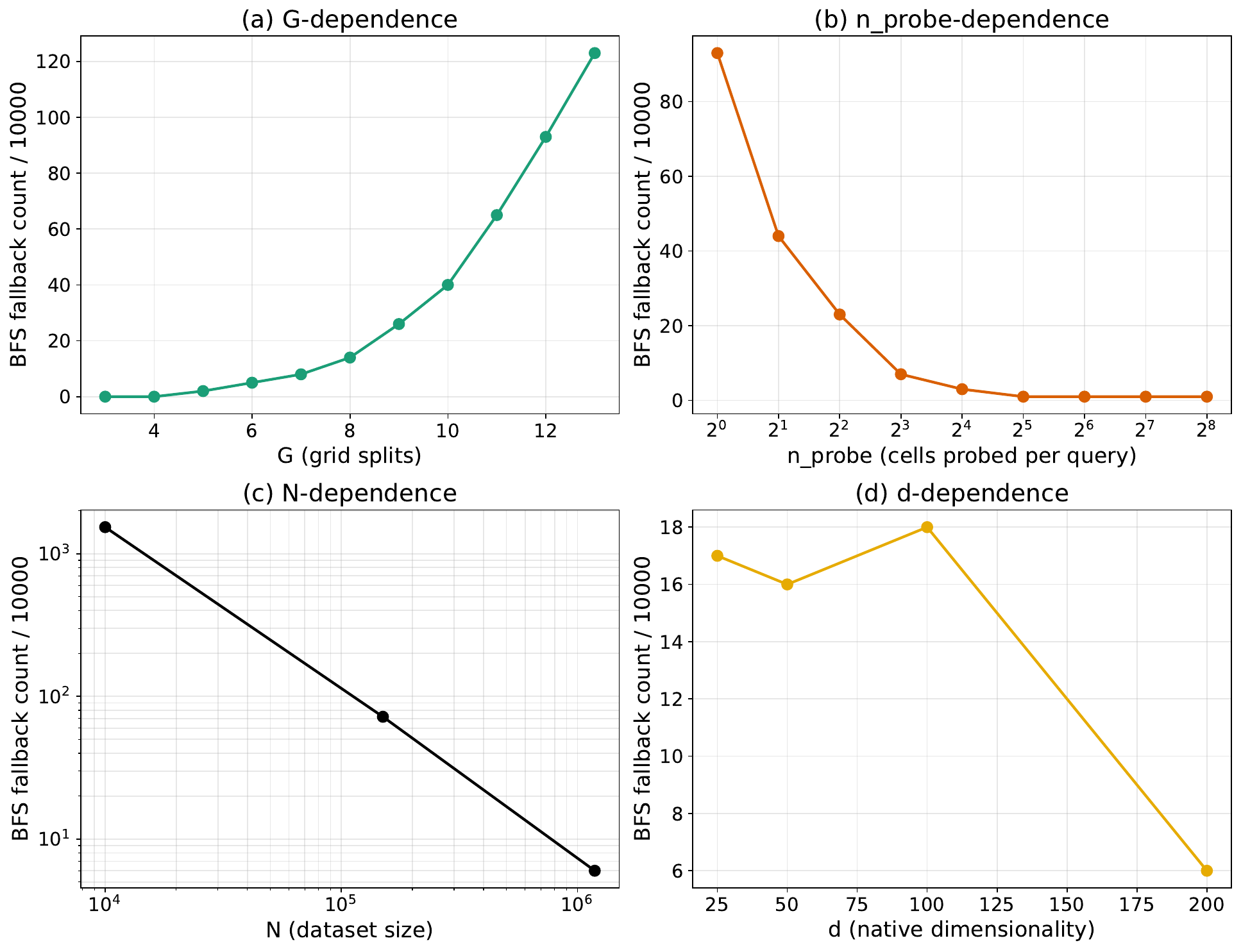}
    \caption{BFS fallback frequency with grid splits ($G$), probe budget ($n_{\text{probe}}$), dataset size ($N$), and dataset dimensionality ($d$). Each panel varies one of the variables while holding the other three constant. Hyperparameter configurations are representative of the range explored during Optuna hyperparameter optimization (Table~\ref{tab:tableE1}). \textbf{(a)} BFS fallback count vs.\ grid splits $G$ on GloVe-200-angular ($N=1.18\times10^6$), with $m=6$ and $n_{\text{probe}}=1$ held fixed. \textbf{(b)} BFS fallback count vs.\ probe budget $n_{\text{probe}}$ on GloVe-200-angular ($N=1.18\times10^6$), with $m=6$ and $G=12$ held fixed. \textbf{(c)} BFS fallback count vs.\ dataset size $N$ on GloVe-200-angular subsampled at $N \in \{10^4, 1.5\times10^5, 1.18\times10^6\}$, with $m=6$, $G=10$, and $n_{\text{probe}}=4$ held fixed. \textbf{(d)} BFS fallback count vs.\ dataset dimensionality $d$ across the GloVe-25, -50, -100, and -200-angular datasets ($N=1.18\times10^6$ each), with $m=6$, $G=10$, and $n_{\text{probe}}=4$ held fixed. }
    \label{fig:figureD12}
\end{figure}
\clearpage

\FloatBarrier

\section{Supplementary Tables} 
\renewcommand{\thetable}{E\arabic{table}}
\setcounter{table}{0}
\begin{table}[h]                                  
\centering                                             
\caption{Pareto-optimal multiprobe grid configurations $(m, G, n_{\text{probe}})$ from benchmark sweeps at each target recall. Dataset: GloVe-25, -50, -100, and -200-angular ($N=1.18\times10^6$). Each table entry represents the Pareto-optimal configuration whose measured recall is closest to the target recall. Parenthetical values indicate the actual measured recall@$k$=10. Note that the entry for $d$=200 at a target recall of 0.95 matches the entry for $d$=200 at a target recall of 0.90, reflecting the algorithm's recall ceiling at that dimensionality.}    

\label{tab:tableE1}                                                                
\small             

\begin{tabular}{c|cccc} 

\toprule                                                                                                                
Recall@$k$=10 & $d$=25 & $d$=50 & $d$=100 & $d$=200 \\     
\midrule  

0.30 & (5,13,2) \textit{(.27)} & (6,9,4) \textit{(.28)} & (5,14,32) \textit{(.30)} & (5,14,32) \textit{(.28)} \\        
0.40 & (5,13,4) \textit{(.40)} & (6,9,8) \textit{(.40)} & (6,9,16) \textit{(.38)} & (5,7,4) \textit{(.38)} \\           
0.50 & (6,9,4) \textit{(.50)} & (5,13,32) \textit{(.50)} & (6,9,32) \textit{(.50)} & (7,7,32) \textit{(.49)} \\         
0.60 & (5,14,16) \textit{(.62)} & (7,7,16) \textit{(.60)} & (8,2,8) \textit{(.61)} & (7,7,64) \textit{(.60)} \\         
0.70 & (5,14,32) \textit{(.71)} & (7,7,32) \textit{(.72)} & (7,2,8) \textit{(.71)} & (8,5,32) \textit{(.68)} \\         
0.80 & (7,7,16) \textit{(.80)} & (7,7,64) \textit{(.81)} & (8,4,32) \textit{(.77)} & (6,5,16) \textit{(.78)} \\         
0.90 & (6,9,64) \textit{(.90)} & (6,5,16) \textit{(.89)} & (6,5,32) \textit{(.90)} & (6,2,4) \textit{(.90)} \\          
0.95 & (2,4,2) \textit{(.95)} & (6,5,32) \textit{(.95)} & (8,4,128) \textit{(.94)} & (6,2,4) \textit{(.90)} \\                                                                                           
\bottomrule                                                                                                             
\end{tabular}                                                                                                           
\end{table}

\end{document}